\documentclass[11pt]{article}

\usepackage[final]{acl}

\usepackage{times}
\usepackage{latexsym}
\usepackage{booktabs}
\usepackage[T1]{fontenc}

\usepackage[utf8]{inputenc}

\usepackage{microtype}

\usepackage{inconsolata}

\usepackage{graphicx}
\usepackage{amsmath}

\usepackage{xcolor}
\usepackage{CJKutf8}
\newcommand{\zh}[1]{\begin{CJK}{UTF8}{gbsn}#1\end{CJK}}
\DeclareRobustCommand{\Rp}{R\ensuremath{^{+}}}
\DeclareRobustCommand{\Rn}{R\ensuremath{^{-}}}
\DeclareRobustCommand{\Rpsub}[1]{R\ensuremath{^{+}_{\text{#1}}}}
\DeclareRobustCommand{\Rnsub}[1]{R\ensuremath{^{-}_{\text{#1}}}}
\newcommand{\repourl}{\url{https://github.com/YuhanXia0809/Knowing-Without-Naming}}

\title{Knowing Without Naming: Availability Without Accessibility in Language Models, and the Tip-of-the-Tongue Talk That Misses It}

\author{
  \textbf{Yuhan Xia\textsuperscript{1}},
  \textbf{Jinming Zhang\textsuperscript{1}},
  \textbf{Yucheng Lin\textsuperscript{2}},
  \textbf{Yunfei Long\textsuperscript{1}},
  \textbf{Massimo Poesio\textsuperscript{3,1}}
\\
  \textsuperscript{1}Queen Mary University of London,
  \textsuperscript{2}University of New South Wales,
  \textsuperscript{3}Utrecht University
\\
  \texttt{\{yuhan.xia, jinming.zhang, yunfei.long, m.poesio\}@qmul.ac.uk}
\\
  \texttt{yucheng.lin@unsw.edu.au}
}

\begin{document}
\maketitle
\begin{abstract}
When a word is on the tip of the tongue, people cannot recall it but can
still recognize it and guess its first letter. Do language models
have an analogous state, and do their own reports reveal it?
We build a bilingual dataset of 2,406 definition--word items and test six open-weight models.
Models persistently fail to name 15--29\% of English and 57--72\% of Chinese items,
yet recognize most of these words in a multiple-choice test.
Our central test is causal: after a model fails 21 times, giving the correct first letter or character
raises production by 50 points in English (30 in Chinese) over a matched retry, while a wrong one pushes it below that retry.
First-letter cues elicit real words but almost never pseudowords, and in English they favor the target over
equally fitting decoys, though our controls cannot separate retrieval from constrained search; Chinese is a boundary case.
We call this graded, \emph{cue-dependent availability}. Models' self-reports do not separate recognition-positive
from recognition-negative failures, whereas signals computed from the model tell failed from successful naming, two of them
before any output. We release the dataset and code.
\end{abstract}

\section{Introduction}
\label{sec:intro}

When a person cannot think of a word, the failure is rarely total. In a
tip-of-the-tongue (TOT) state \citep{brown1966tip,brown1991tot}, the
speaker cannot say the word but can still reject wrong candidates,
guess its first letter, and often say roughly how long it is. The
speaker also knows, correctly, that the word would be recognized on
sight, and can work around the gap by describing it or by switching
languages. Theories of speech production are built to explain this
partial access \citep{levelt1999theory,dell1986spreading}, and a
feeling-of-knowing layer \citep{hart1965memory,nelson1990metamemory}
keeps speakers calibrated to their own retrieval states.

Language models behave differently. Asked for a word they cannot
retrieve, they rarely pause, describe it, or hedge. In contrast, they
answer, and the answer looks fluent and confident but is always
problematic. Whether anything like a TOT state lies behind that
problematic answer raises two questions. First, do knowing and naming come
apart in models the way they do in people? If they do, not every naming
failure is ignorance. Some would be failures to reach a word the model
has, which is a different problem to diagnose and to repair. Second, does the state have a signature, in the model's own reports or in signals computed from it, that a system could catch
before it commits to a name?

TOT supplies our paradigm, not our theory. From it we take probing
inside the failure episode after a truthful rejection and the contrast
between a correct and a wrong first unit; we invert its criterion,
using recognition rather than self-report (\S\ref{sec:construct}). What
we test in models is availability without accessibility in Tulving's
sense \citep{tulving1966availability}, not a human TOT state.

One episode shows the pattern. Asked for the word defined as ``the
belief that countries should manage their international relations
independently, without seeking input or collaboration from other
countries,'' Qwen2.5-72B answers \emph{isolationism} in all 21
attempts. It still picks \emph{unilateralism} out of four options in
all four orderings we test. Told in addition that the word begins with
\emph{u}, it produces \emph{unilateralism} in 20 of 20 samples; a
fresh attempt without the letter gives 0 of 20, and so does an attempt
with a wrong letter. Given the option of reporting a tip-of-the-tongue
state, it answers \emph{isolationism} again. Given the same options
inside the failed episode, after its answer has been rejected, it
replies that it does not know the word. This episode is unusually
clean; the rest of the paper measures how often, and how far, the
pattern holds.

In NLP, work on TOT has been about helping a person who is
already in the state \citep{arguello2023trec,bhargav2022tot}. Whether a model is ever in anything like one has not, to our
knowledge, been tested. Testing it needs a dataset nobody had built: items designed to make
models fail at naming, with a single
definition shared by every task that uses one, and an answer key clean
enough that a failure means what it looks like.

We take up the first question, whether knowing and naming dissociate,
with controls that never rely on what the model says about itself. Following \citet{brown1966tip}, we define an
\emph{availability state}: the word is \emph{available}, in that the
model recognizes it and a small hint about its form brings it out, but
not \emph{accessible}, since the model cannot produce it on demand.
Operationally, a persistent naming failure that the model still
recognizes is a \emph{recognition-positive failure}, and one it does not
recognize is a \emph{recognition-negative failure}; we keep ``TOT'' for
the human state and for what the models themselves report.

The difficulty is that recognition on its own cannot establish
availability. A
definition can point clearly enough at the answer that a model could
pick it out of four options without ever having stored the word. So we
use recognition only as a filter, and put the weight on production.
After a model has failed to name the target, we give it the target's
first letter or character and see whether the word now comes out. That test is generative and within-item, so the usual objections to
multiple choice do not reach it. It works, and by a wide margin. Each further control bounds one
competing explanation: that any hint
would have worked as well, that a harder recognition test would make
the state disappear, and that the model is \emph{building} the word
out of the hint and the definition rather than retrieving one it
already had. The dissociation holds up in English. In Chinese it
survives only at the edge of what we can detect once the harshest
filters are applied, which we report as a boundary rather than drop.

On the second question the channels separate, though not where one
might expect. Neither the models' own reports nor the signals computed
from them reliably say which kind of failure it will be, and how much a
model reports at all is a trait of its family rather than a reading of
its state. What they do tell apart is naming failure from success.

We make four contributions. (1)~Availability without accessibility,
established causally: the correct first unit raises recovery by 50.3 percentage
points in English and 29.8 in Chinese over a matched retry that
withholds only the cue, and a \emph{wrong} first unit pushes production
below that retry, so the effect is form-specific
(\S\ref{sec:construct}). (2)~Scale and generality: among failures
with a clean recognition verdict, recognition-positive failures outnumber
recognition-negative ones 11.2:1 in English and 4.9:1 in Chinese, and the
taxonomy replicates across fourteen models from seven lineages
(\S\ref{sec:taxonomy}). (3)~A state--report dissociation with a
practical consequence: every signal computed from the model separates
failed from successful naming and two are readable before any output
token, while the model's verbalized confidence sits at chance and its
self-reported TOT does not separate the two failure types. A system that
needs to know when a name is unsafe should measure the model rather than
ask it (\S\ref{sec:mimicry}--\ref{sec:probe}). (4)~The dataset and
code: 2,406 bilingual, leakage-filtered items with vetted answer sets
driving all six tasks\footnote{Code and data: \repourl}
(Appendix~\ref{app:battery}). Building them yielded a methodological finding
of independent value: a metalinguistic probe asked in a fresh context
drifts to whichever candidate the definition brings to mind
(\emph{referent drift}), so such probes have to be asked inside the
failure episode.

\section{Related Work}
\label{sec:related}

\paragraph{The human state, and what it took to establish one.}
TOT was established by asking people what they could report about a
word they could not say \citep{brown1966tip}, and the paradigm has
been refined since \citep{brown2012tip}. Two of its
findings shape our design. Hart's feeling-of-knowing paradigm and TOT
proper are distinct, because TOT episodes are a minority of
recognition-positive recall failures
\citep{schwartz2011tot,schwartz1997tot}; our labels inherit that
distinction and quantify it. And proper names behave unlike common
ones \citep{cohen1986names}, which is why we filter them out. We take
the in-episode probing protocol from \citet{brown1966tip}, the
transmission-deficit account from \citet{burke1991tip}, and the
practice of porting psycholinguistic paradigms to models from
\citet{linzen2016assessing} \citep{xia2024sensoryt5,zhao2025leveraging,xia2024sensory}. The bilingual TOT literature
\citep{gollan2001tot,gollan2004tot,pyers2009bimodal} motivates the
Chinese half of the dataset, where a first \emph{character} carries more
than a first letter does.

\paragraph{NLP has studied the user's state, not the model's own.}
A retrieval literature resolves human TOT queries
\citep{arguello2023trec,bhargav2022tot,chwang2025blur,he2026multitot}.
\citet{xu2024tip} use the same elicitation task
\citep{hill2016reversedict,zhang2020multird,qi2020wantwords,
mickus2022codwoe,almeman2023dex} as a capability probe and analyze the successes, whereas we classify and
validate the \emph{failures}.
\citet{balepur2025reverse} show reverse-QA errors track
answer-frequency in pretraining data
\citep{kandpal2023longtail,mallen2023popqa}. To our knowledge no existing resource is built to elicit and
behaviorally label a model's own lexical retrieval failures, which is
why the dataset has to come before the analysis.

\paragraph{Hidden knowledge and recall bottlenecks.}
Models encode more than they emit
\citep{azaria2023internal,orgad2025know,gekhman2025inside,
ferrando2025know}. For parametric facts the bottleneck is recall
rather than encoding \citep{calderon2026empty}, and some
hallucinations occur while the correct answer still carries mass
\citep{yeom2026commitment}. We presuppose that bottleneck and add
the \emph{lexical} characterization, which enables a measurement a factual-QA design cannot express,
namely form-specific causal cueing. Our cue probes are the
lexical, causal analogue of the test-time reasoning that partially
rescues recall \citep{gekhman2026thinking}.

\paragraph{Metacognition and calibration.}
P(IK) self-evaluation \citep{kadavath2022know}, semantic-entropy
uncertainty \citep{farquhar2024detecting}, metamemory comparisons
\citep{huff2025judgments}, and confidence elicitation for compute
gating \citep{cao2026know} establish that some calibration signal
exists. Against this, a skeptical literature finds model introspective reports to be post-hoc narration
\citep{singh2026introspect,ackerman2026limited}. Our feeling-of-knowing (FOK) probe is elicited before any retrieval
attempt, as a cue-familiarity judgment \citep{reder1992what}.

\section{Dataset, Tasks, and State Labels}
\label{sec:method}

\subsection{Building the dataset}
\label{sec:items}

A model can fail to name a word for reasons unrelated to retrieval.
The definition may fit several words. The answer may be a proper noun
the model was never expected to know. The definition may be a
dictionary entry the model has seen verbatim. The dataset is built to
screen these out or reduce them before the experiment starts.

English definitions come from WordNet noun glosses
\citep{fellbaum1998wordnet}. Chinese ones come from the moedict
dictionary \citep{moedict2015}, converted to simplified characters with
OpenCC. From each source we drew two pools, one normed on SUBTLEX
frequency and one reaching beyond that frequency corpus
\citep{speer2022wordfreq}. We then filtered out proper nouns and any
item whose definition leaks the answer's morphology. That leaves 2,406
frequency-stratified definition--word items, 1,503 English and 903
Chinese.

Qwen2.5-72B, the judge, then annotates every surviving item three
ways. It collects the set of answers that should count as correct. It
assesses whether the definition picks out one word rather than several
(uniqueness). And it writes a paraphrase that avoids naming the answer.
Wherever a task shows a definition, the model sees this paraphrase and
not the original. So those tasks all work from a single stimulus, and
no prompt matches a dictionary entry word for word. Paraphrasing could
in principle break uniqueness, so we re-ran the uniqueness check on the
paraphrase itself. The two verdicts agree on 91.5\% of English and
88.6\% of Chinese items (Appendix~\ref{app:battery}).

Two checks bound the judge. A second-family judge,
Llama-3.1-70B, agrees with the Qwen uniqueness verdicts on 90.0\% of a
400-item English sample and with the Qwen reversal verdicts at
\(\kappa{=}0.76\)--0.83. And three graduate annotators audited 210
items. Their majority matches the judge on 92--96\% of synonym,
paraphrase and reversal decisions but on 66\% of uniqueness ones, so we
re-test the central result on human-checked items
(\S\ref{sec:construct}).

\subsection{What each model is asked to do}
\label{sec:stages}

Each model runs the six tasks below on every item, each task in a
fresh context unless noted. The prompts are reproduced in
Appendix~\ref{app:prompts}.

\textbf{S1, production.} Name the word the definition describes. We
take one greedy decode and 20 samples at \(T{=}1.0\). A \emph{strict
failure} means no acceptable answer appears, in the greedy decode or in
any of the 20 samples.

\textbf{S2, feeling of knowing.} Before any attempt, the model rates
from 0 to 100 how confident it is that it could pick the word out of
four options. We separately ask the same thing as a yes/no question
and read the probability it puts on the token \emph{Yes}.

\textbf{S3, recognition.} A four-alternative forced choice (4AFC): the
model must pick one of four options and has no way to decline. The
three distractors are ours: an orthographic neighbor (a word close in
written form) where the pool has one, an associate that overlaps the
definition, and a random word matched for frequency and length
(selection procedure in Appendix~\ref{app:battery}). The options are
rotated through all four positions, and success means at least 3 of
the 4 rotations correct.

\textbf{S4, partial access, asked inside the episode.} A probe such as
\emph{``what letter does it begin with?''} answers about whatever word
the definition brought to mind, which need not be the target
(\emph{referent drift}). We therefore ask it inside the production
episode, after the model's wrong answer has been truthfully rejected
\citep{brown1966tip}.

\textbf{S5, self-report.} The model may answer with the word, or reply
\emph{TOT} if it feels it knows the word but cannot retrieve it, or
\emph{DK} if it does not know it. This never enters the state labels.

\textbf{S6, reversal control.} Given the word, the model defines it.
Every model does this. It is the separate judging of those
definitions, not the task, that runs on fewer models
(\S\ref{sec:stats}).

\subsection{Turning failures into two labels}
\label{sec:labels}

A person gets one attempt at a word. A sampling model gets as many as
we allow, so the threshold must be chosen rather than inherited. We
use 21 attempts: one greedy decode plus 20 samples at \(T{=}1.0\).
Twenty is a practical compromise. A word the model would emit on 10\%
of draws escapes 10 samples 35\% of the time but 20 samples only
12\%, and one it would emit on 15\% escapes 20 samples only 4\%. Zero
hits still do not prove zero support: the one-sided 95\% bound on the
per-draw probability is about 14\% (rule of three).

Write \(k_{\text{prod}}\) for how many of the 20 samples hold an
acceptable answer. We label an item \textbf{known} if
\(k_{\text{prod}}\geq6\). It is a \textbf{recognition-positive failure}
(\Rp{}) if production fails strictly and recognition succeeds, and a
\textbf{recognition-negative failure} (\Rn{}) if production fails
strictly and recognition fails too, meaning at most 1 of the 4
rotations correct. \Rp{}:\Rn{} is their ratio. Items in between are
left out of the binary contrasts: 2 of 4 in recognition, the 1--2 and
3--5 bands of \(k_{\text{prod}}\), and the few items with no scoreable
recognition record. Counts for each model and language ship with the
release. Neither the feeling-of-knowing ratings nor the in-episode
probes enter these definitions, so nothing the model says about itself
can make the labels circular.

What a larger budget would buy we can measure. The \emph{matched
retry} of \S\ref{sec:construct} reruns the same prompt on the same
items, without a cue, for 20 more samples. Doubling the budget that
way recovers the target on 11.2\% of the classified English failures
(the \Rp{} and \Rn{} items) and 4.5\% of the Chinese ones. It would
come at twice the cost for every later production condition, all of
which reuse this budget so that the comparisons stay matched. We
therefore read a strict failure as persistent non-production at this
budget.

Strict failure depends on decoding as well as on budget. Rerunning
production for the two smallest models at \(T{=}0.7\) or at top-\(p\) 0.9
adds failures but leaves the \Rp{}:\Rn{} ratio and the cue effect
similar, and across budgets the ratio flattens after ten samples
(Appendix~\ref{app:battery}, Table~\ref{tab:decoding}).

\subsection{Models and statistics}
\label{sec:stats}
We run six models in two languages:
Qwen2.5-7B/14B/32B/72B \citep{qwen2025qwen25} and Llama-3.1-8B/70B
\citep{grattafiori2024llama3}, under vLLM \citep{kwon2023vllm}. One
model in one language is a \emph{setting}, so there are twelve
settings in all. Every main-text number is over these unless we say
otherwise. A few additional analyses run on the smallest and largest
of each family, Qwen-7B/72B and Llama-8B/70B, covering both ends of
the scale range; Appendix~\ref{app:battery} lists which. Everything
else, including the six tasks and the adversarial recognition of
\S\ref{sec:bounds}, runs on all six.

A post-hoc \emph{sweep} of fourteen models across seven lineages
checks the state labels and the report-style result beyond two
families (Appendix~\ref{app:taxdetail}). Every effect estimate and
validity test in the main text uses the six, except the two that use
the sweep: the replication of the taxonomy (\S\ref{sec:taxonomy}) and
the fresh-context self-report analyses (\S\ref{sec:mimicry}).

All inferential summaries are exploratory, with descriptive
\(p\)-values. Because the dataset is shared, one item contributes up
to six correlated rows, and a binomial interval would be far too
narrow. Every interval on a pooled ratio is therefore an
\emph{item-cluster} bootstrap percentile interval over 1,000
resamples, with each model's counts recomputed inside the replicate
(Appendix~\ref{app:battery}). Per-experiment coverage is stated where
each experiment is reported.

\section{A Frequent Behavioral State}
\label{sec:taxonomy}
\begin{table}[t]
\centering
\scriptsize
\setlength{\tabcolsep}{2.9pt}
\begin{tabular}{llrrrrr}
\toprule
Lang & Model & Fail\% & \Rp{} & \Rn{} & Ceil\% & \Rp{}:\Rn{} \\
\midrule
EN & Qwen2.5-7B & 29.4 & 395 & 30 & 97.8& 13.2 \\
EN & Qwen2.5-14B & 24.7 & 331 & 26 & 97.9& 12.7 \\
EN & Qwen2.5-32B & 23.3 & 321 & 20 & 98.8& 16.1 \\
EN & Qwen2.5-72B & 18.7 & 253 & 18 & 98.5& 14.1 \\
EN & Llama-3.1-8B & 23.3 & 292 & 41 & 97.9& 7.1 \\
EN & Llama-3.1-70B & 14.5 & 187 & 24 & 98.2& 7.8 \\
\midrule
ZH & Qwen2.5-7B & 71.7 & 513 & 95 & 94.8& 5.4 \\
ZH & Qwen2.5-14B & 64.5 & 469 & 80 & 94.6& 5.9 \\
ZH & Qwen2.5-32B & 61.0 & 469 & 63 & 95.1& 7.4 \\
ZH & Qwen2.5-72B & 57.4 & 423 & 75 & 93.3& 5.6 \\
ZH & Llama-3.1-8B & 69.1 & 429 & 137 & 87.2& 3.1 \\
ZH & Llama-3.1-70B & 58.0 & 386 & 101 & 91.8& 3.8 \\
\bottomrule
\end{tabular}
\caption{State labels for the six models. \emph{Fail\%} =
strict production failures (greedy wrong and 0/20) as a percentage of
that language's dataset (1,503 En / 903 Zh). \emph{\Rp{}} / \emph{\Rn{}} =
those failures with $k_{\mathrm{rec}}\!\geq\!3/4$ / $\leq\!1/4$.
\emph{Ceil\%} = recognition on the model's own \emph{known} items
($k_{\mathrm{prod}}\!\geq\!6/20$), the instrument check, all above
the 85\% gate. \emph{\Rp{}:\Rn{}} = their ratio. \Rp{} and \Rn{} do not sum to the
failure count, because the $k_{\mathrm{rec}}\!=\!2/4$ band and
unresolvable items are excluded from binary contrasts; those counts
ship with the release. Pooled intervals are in \S\ref{sec:taxonomy},
the fourteen-model version in Table~\ref{tab:sweep}.}
\label{tab:taxonomy}
\end{table}

Models recognize most of the words they consistently fail to produce.
A strict production failure is a word the model does not produce in
any of 21 attempts. Such failures cover 15--29\% of the English
dataset and 57--72\% of the Chinese one, which was built to be harder.
The four-option recognition test then sorts them. A
recognition-positive failure (\Rp{}) is a word the model still picks
out in at least 3 of the 4 orderings of the options, and a
recognition-negative failure (\Rn{}) one it picks out in at most 1.
Among failures with a clean recognition verdict, recognition-positive
failures outnumber recognition-negative ones 11.2:1 [8.6,\,15.8] in
English and 4.9:1 [4.2,\,5.8] in Chinese (Table~\ref{tab:taxonomy},
Figure~\ref{fig:space}).

Two checks say this is not an instrument artifact. First, the judge
gives each item a verdict on whether its definition picks out one
word, its uniqueness certificate. Re-drawing each item's certificate
at the judge's measured error rate widens the intervals without
bringing them near parity. Second,
recognition on each model's own known items, those it produces in at
least 6 of 20 samples, clears the 85\% gate in every setting
(Appendix~\ref{app:battery}, Appendix~\ref{app:taxdetail}).

Nor is it an artifact of scale. Larger models fail less often, while
the ratio moves non-monotonically, and the taxonomy replicates across
fourteen models from seven lineages (Table~\ref{tab:sweep}). The ratio
moves with the threshold, but no single criterion inverts it, and
floor corrections only compress it (\S\ref{sec:bounds},
Appendix~\ref{app:taxdetail}). The nearest human anchor, four-option
recognition of \emph{all} unrecalled general-knowledge answers, runs
at 33.7\% against 25\% chance \citep{norman2016fok}, far below what we
see here. That distance is why we claim an availability state rather
than TOT proper.

\paragraph{Why the forced choice cannot settle it.} A four-option test
can be passed by inferring which option fits the definition, without
stored knowledge of the word, so we measure that route directly. We
replace the target with the item's strongest definition-compatible
non-answer and leave it among the item's own three distractors. Models
still choose this decoy often, in 27.6--41.2\% of cases. What
discriminates is the \emph{margin}: with the target restored, the same
settings choose the target 48--61 points more often than they chose
the decoy. Because the decoy is a
judged non-answer, that margin is at best an \emph{upper} bound on the
target-specific part of recognition. Decoy endorsement does not fall
at 70B/72B (Table~\ref{tab:validity}, Appendix~\ref{app:taxdetail}).

\section{Cue-Unlocked Production}
\label{sec:construct}
\begin{table}[t]
\centering
\scriptsize
\setlength{\tabcolsep}{3pt}
\begin{tabular}{@{}lrr@{}}
\toprule
 & English & Chinese \\
\midrule
\multicolumn{3}{@{}l@{}}{\textbf{The challenge}: a 4AFC can be passed by definition inference}\\
\quad Decoy endorsed, target deleted & 29.0--41.2\% & 27.6--35.7\%\\
\quad\quad \emph{margin to target-present} & 48--61 pts & 48--54 pts\\
\midrule
\multicolumn{3}{@{}l@{}}{\textbf{Test~1} adversarial 4AFC: \emph{bounds distractor easiness}}\\
\quad \Rp{}:\Rn{} & 5.7:1 & 3.2:1\\
\midrule
\multicolumn{3}{@{}l@{}}{\textbf{Test~2} \(+\)reversal: \emph{a filter, not a certification}}\\
\quad \Rp{}:\Rn{}, per setting & 5.2--10.8:1 & 1.2--3.7:1\\
\quad Composed with Test~1 & 3.90:1 & 1.66:1\\
\midrule
\multicolumn{3}{@{}l@{}}{\textbf{Test~3} cue-unlocked production: \emph{needs no option set}}\\
\quad Matched retry \(\to\) correct first unit & 11.2\%\(\to\)61.6\% & 4.5\%\(\to\)34.3\%\\
\quad \textbf{Increment} (the estimand) & \textbf{+50.3 pts} & \textbf{+29.8 pts}\\
\quad Wrong first unit & 4.8\% & 1.3\%\\
\midrule
\multicolumn{3}{@{}l@{}}{\quad\emph{The same cue, aimed at three things}: \emph{what it can reach}}\\
\quad Pseudoword first unit & 0\% & 0--0.3\%\\
\quad Stored decoy first unit & 14--21\% & 11--29\%\\
\quad The target, same items & 53--73\% & 20--46\%\\
\bottomrule
\end{tabular}
\caption{What each of the three tests bounds. Test~1 runs on all six models; the reversal \emph{judging} in Test~2 and
the pseudoword--decoy--target comparison run on four; Test~3 runs on all twelve
model\(\times\)language settings; that comparison runs on one
matched population (English settings 60--128 items). \emph{Margin to target-present} = how far 4AFC accuracy rises when the
deleted target is restored to the same option set, against the
decoy-endorsement rate above it. Decoy-to-target margins are
2.0--2.9\(\times\) (En) / 1.6--2.0\(\times\) (Zh) under
scoring-matched credit, and 3.1--4.2\(\times\) / 1.6--2.1\(\times\)
unrestricted. Increments are computed on unrounded condition rates, so they
need not equal the difference of the rounded rates above them; the two rates around the arrow count the 20 samples of each arm, while the other cue rates also count the greedy decode after the cue.
Per-setting values, inference controls A--D and Wilson intervals:
Appendix~\ref{app:taxdetail}, Appendix~\ref{app:sigdetail}.}
\label{tab:validity}
\end{table}

This section tests Hart-style \emph{availability without accessibility}
\citep{tulving1966availability,hart1965memory}. The human literature
makes the \emph{report} the criterion and uses recognition to validate
it \citep{koriat1974tot,gollan2006tot}. Because the report's validity
is our question, we swap the two. Recognition can be passed by
inference (\S\ref{sec:taxonomy}), so the test here uses no options
(Test~3 in Table~\ref{tab:validity}). After a model has failed to
produce a word, we give it one unit of the word's form, its first
letter in English or its first character in Chinese, and ask again.
The design is causal and within-item, like the canonical human test
\citep{brown1991tot,james2000tot}. \S\ref{sec:bounds} then tests the
recognition filter itself.

\paragraph{What the cue is compared with.} Every item in this test is
one the model failed on in all 21 attempts, so its uncued production
is zero by construction. Zero hits in 21 attempts do not prove that the
word can never come out (\S\ref{sec:labels}), so the baseline is a
second try. We rerun the same prompt on the same items, without a cue,
with a new seed and again 20 samples at \(T{=}1\), and call this
the \emph{matched retry}. The cued run adds the first unit and changes
nothing else. A third run supplies a wrong first unit. All runs use
the same 5,178 classified failures.

\begin{figure}[t]
\centering
\includegraphics[width=\columnwidth]{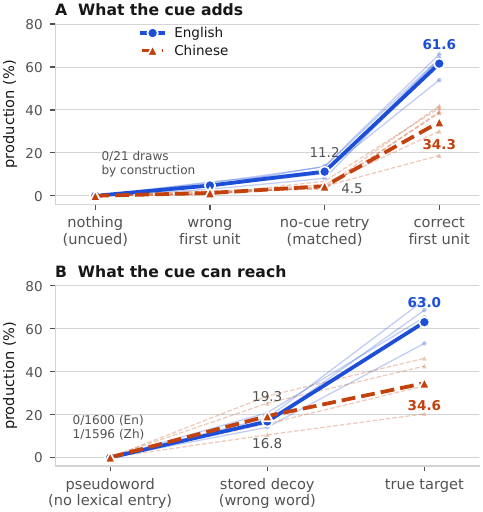}
\caption{The cue test, on cleanly classified failures.
\textbf{(A)} Production after a failed attempt, in three conditions:
the matched retry without a cue, the correct first unit, and a wrong
first unit. Thin lines are the six models per language, thick lines
the pool; uncued production is 0 across 21 attempts by construction.
\textbf{(B)} The same cue aimed at a pseudoword, at a stored decoy, and
at the true target on the same items, on one matched population, for
four models per language (the smallest and largest of each family).
Panel B plots unrestricted rates; the margin we carry is the
scoring-matched one (\S\ref{sec:construct}). Panel A counts the 20
samples of each arm, as the increment does; panel B and the
per-setting values in Table~\ref{tab:cue} also count the greedy decode
after the cue.}
\label{fig:anchor}
\end{figure}

\paragraph{The cue recovers the word.} With the correct first unit,
production rises over the matched retry by \(+50.3\) points in English
and \(+29.8\) in Chinese (Table~\ref{tab:validity},
Figure~\ref{fig:anchor}A). The increment is positive in each of the 12
settings, with every interval excluding zero
(Appendix~\ref{app:sigdetail}; Table~\ref{tab:cue} gives the rates by
setting). A \emph{wrong} first unit does the opposite: it pushes production below
the matched retry, so the effect is form-specific. Human studies of
blocking cues report weaker interference
\citep{jones1987blocking,meyer1992bock}. The drop is still far smaller
than what the correct unit adds, so we measure the cue by its
increment over the matched retry, not by the difference between the
correct and the wrong unit (Appendix~\ref{app:sigdetail}).

\paragraph{What the cue reaches.} We next aim the same kind of cue at
three different things, on four models per language
(Figure~\ref{fig:anchor}B). The first is a \emph{pseudoword}, a form
built to match no lexical entry. Given its first unit, the models
produce it in 0 of 1,600 English cases and 1 of 1,596 Chinese ones.
The second is a \emph{decoy}, a real word that is not the answer to
the definition. Given the decoy's first unit, the models do produce
the decoy on some items. The third is the true target on those same
items. It comes out 2.0--2.9 times as often as the decoy in English
and 1.6--2.0 times as often in Chinese, when both are scored the same
way on single-answer items (Appendix~\ref{app:sigdetail}).

\paragraph{Retrieval or constrained search?} These comparisons show
what a cue can \emph{reach}: stored real words, and almost never the
tested pseudowords. They do not show how the word is reached. The
model may retrieve the stored word, or it may compose the answer by a
constrained search, looking among stored words for one that fits both
the definition and the cue. Because the prompt asks for a real word,
such a search predicts the pseudoword result as well as retrieval
does, and it also predicts the drop under a wrong first unit, so
neither separates the two accounts. One comparison is harder for
constrained search to explain: a preference for the target over a
decoy that fits the definition equally well. At matched answer-level definition fit, the
English target is still reached on 28.9 points more of the same source
items than the decoy. This rests on few items and a wide interval, and
the Chinese margin is null, so we limit this reading to English. The
English preference persists at matched fit when the comparison is
restricted to first letters cued in both arms, and across all
classified English failures the unlock rate varies by only about 9
points between terciles of cohort size, the number of words that share
the first letter (Appendix~\ref{app:sigdetail}).

\paragraph{Chain of thought.} Asking the model to think step by step,
without any cue, also recovers words. It was run with five samples;
over the matched retry cut to the same five, it adds \(+31.1\) points
in English and \(+5.8\) in Chinese. It also changes the instruction
and the output format, so the matched retry stays the control
(Appendix~\ref{app:sigdetail}).

\paragraph{Items judged uniquely answerable.} A second reading is that
the cue only picks one of several defensible answers. Restricting to
English items that two judges independently call uniquely answerable
leaves the increment at \(+53.3\) points. Three annotators then
checked every English failure item by hand. A majority confirms 330 of
the 640 classified ones as uniquely answerable, and on these the
increment is \(+50.5\) points (Appendix~\ref{app:sigdetail}).

\paragraph{Graded, and domain-specific.} Availability is graded across
failure strata. Under the correct first unit, recognition-positive
(\Rp{}) failures unlock more often than recognition-negative (\Rn{})
ones in all 12 settings (Appendix~\ref{app:sigdetail}), and \Rn{}
items unlock too. The claim is
therefore that knowledge is present \emph{in failures} and
cue-reachable, not that \Rp{} failures have a property \Rn{} failures
lack. We read ``availability'' operationally, in Tulving's
cue-dependent sense: graded sub-threshold accessibility that a form
constraint amplifies \emph{is} availability on this reading. The test
is also domain-specific. Run unchanged on 300 non-lexical facts, it
does not fire: it unlocks 2.0\% where the same intervention reaches
63\% on English words (Appendix~\ref{app:discdetail}).

\section{Bounding the Alternatives}
\label{sec:bounds}
Section~\ref{sec:taxonomy} counted two kinds of failure:
recognition-positive failures, words a model cannot produce but still
recognizes, and recognition-negative ones, which it can neither
produce nor recognize. The first outnumber the second by 11.2:1 in
English and 4.9:1 in Chinese. That ratio (\Rp{}:\Rn{}) depends on how
strict the recognition test is, so we repeat it under two stricter
tests, each bounding a different alternative. \textbf{Test~1},
adversarial recognition, uses harder wrong options: all three
distractors are rebuilt from the highest definition-overlap neighbors.
It checks whether the models passed recognition only because the
original distractors were easy; it does not rule out inference.
\textbf{Test~2} adds a second requirement: shown the word, the model
must also define it correctly (a word\(\rightarrow\)definition
reversal). That is a stricter \emph{filter}, though passing it still
does not establish that the word's \emph{form} is available. A third
alternative, reverse-direction knowledge, does not account for the
increment; the last paragraph of this section shows why.

\begin{table}[t]
\centering
\scriptsize
\setlength{\tabcolsep}{2.6pt}
\begin{tabular}{llrrrr}
\toprule
Lang & Model & \Rpsub{joint} & vs.\ \Rnsub{adv} & vs.\ \Rnsub{joint} & adv.\ only \\
\midrule
EN & Qwen2.5-7B & 298 & 5.14 & 2.07 & 6.5 \\
EN & Qwen2.5-72B & 175 & 4.17 & 1.65 & 5.5 \\
EN & Llama-3.1-8B & 196 & 2.84 & 1.27 & 3.8 \\
EN & Llama-3.1-70B & 143 & 3.67 & 1.91 & 4.5 \\
\midrule
ZH & Qwen2.5-7B & 312 & 2.05 & 0.93$^\ddagger$ & 3.1 \\
ZH & Qwen2.5-72B & 262 & 2.43 & 1.02 & 3.6 \\
ZH & Llama-3.1-8B$^\dagger$ & 162 & 0.88$^\ddagger$ & 0.35$^\ddagger$ & 2.1 \\
ZH & Llama-3.1-70B & 179 & 1.67 & 0.52$^\ddagger$ & 3.5 \\
\midrule
EN & pooled (4) & 812 & \textbf{3.90} & 1.70 & 5.0 \\
ZH & pooled (4) & 915 & \textbf{1.66} & 0.65$^\ddagger$ & 2.9 \\
\bottomrule
\end{tabular}
\caption{The joint criterion, on the four models whose reversals were
judged. \emph{\Rpsub{joint}}: strict production failures that pass
both adversarial recognition and the reversal, in items. The other
three columns are \Rp{}:\Rn{} ratios that differ only in what counts
as \Rn{}. \emph{vs.\ \Rnsub{adv}}: \Rn{} by adversarial recognition
alone (\(k_{\text{rec}}\leq1\)), the same branch as Test~1; this is the
comparable denominator. \emph{vs.\ \Rnsub{joint}}: \Rn{} is any item
failing \emph{either} control, which also absorbs the ambiguous
\(k_{\text{rec}}{=}2\) band and every control-only item; this is the
literal denominator, and the harsher one. \emph{adv.\ only}: the ratio
under Test~1 alone, without the reversal. \(^\ddagger\) = inverted
(\(<\)1:1). \(^\dagger\) = this setting's \emph{reversal} instrument
fails the 85\% known-item ceiling, so its ratios are
instrument-invalid and its interval covers parity in every reading.
The baselines for this table are the standard \Rp{}:\Rn{} ratios
re-pooled over these four models, 10.0:1 (En) and 4.3:1 (Zh), not the
pooled ratios of \S\ref{sec:taxonomy}. The instrument gate, the
coverage accounting and every per-setting interval are in
Appendix~\ref{app:taxdetail}.}
\label{tab:joint}
\end{table}
\paragraph{The two tests composed.} Under each test alone,
recognition-positive failures still outnumber recognition-negative
ones in both languages (Table~\ref{tab:validity}). When a word must
pass both tests to count as recognized, the ratio shrinks to 3.90:1 in
English and 1.66:1 in Chinese (Table~\ref{tab:joint}). For the same
four models the standard ratio is 10.0:1 and 4.3:1. The two tests give
a ratio and the cue test of \S\ref{sec:construct} an absolute rate, so
they do not measure one thing.

\paragraph{A harsher count.} The two composed ratios count a word as
recognition-negative if it fails adversarial recognition, as Test~1
does. A harsher, intentionally punitive count treats a word as
recognition-negative if it fails \emph{either} test, or if its
recognition result is ambiguous. The English ratio then falls to
1.70:1 and the Chinese one to 0.65:1; three of the four Chinese
settings also fall below 1:1. The Chinese figure is an inversion,
which we report rather than set aside. What survives in English is
therefore compression, not inversion, and the Chinese joint result is
a boundary condition (Table~\ref{tab:joint},
Appendix~\ref{app:taxdetail}).

\paragraph{The reverse-direction artifact.} The third alternative is
that the dissociation reduces to \emph{reverse-direction knowledge}, a
distinct sense from the one in our title: the model can define a word
it is shown and still not name it from a definition. The models'
definitions do get less accurate from known items to
recognition-positive failures to recognition-negative ones, and their
accuracy predicts state, consistent with the reversal curse
\citep{berglund2024reversal}. But this cannot explain the cue result.
In the cue test the correct first unit, the wrong one and the matched
retry are compared on the same item (\S\ref{sec:construct}), so no
property of the item, reverse-direction knowledge included, can
produce the increment
(Table~\ref{tab:validity}). Adding the item's reversal correctness as
a covariate leaves the cue effect intact
(Appendix~\ref{app:taxdetail}).

\section{State-Blind Self-Reports}
\label{sec:mimicry}
\paragraph{The question.} In the self-report task (S5), a model may
reply with the word, with the label TOT if it feels it knows the word
but cannot retrieve it, or with DK if it does not know the word. Raw
agreement between a self-reported TOT and a recognition-positive
failure is bounded by how often a model reports at all, so the
question we ask is base-rate-free: among strict failures, does a
self-reported TOT distinguish recognition-positive (\Rp{}) from
recognition-negative (\Rn{}) failures? We measure this with an odds
ratio; a value of 1 means a TOT report is equally likely for the two
kinds of failure. We ask in two ways. In a \emph{fresh context}, the
question comes in a new conversation. \emph{In-episode}, it comes
inside the failed production episode, after the model's wrong answer
has been rejected, so that the report is about the intended word.

\paragraph{No reliable discrimination is detectable.} The experimental
results show that a self-reported TOT does not distinguish the two
kinds of failure. Asked in-episode, the pooled odds ratio is 1.06
[0.75,\,1.50], and no single setting is significant on its own. Asked
in fresh context, we pool the \emph{live} subset: the 16 of the 28
settings of the fourteen-model sweep (Table~\ref{tab:sweep}) whose
report rate among strict failures falls between 5\% and 90\%. Its
odds ratio is 0.90 [0.74,\,1.09]. We keep those two bounds apart
rather than pooling
them. The preferred in-episode instrument admits positive
discrimination up to 1.50, while the tighter 1.09 ceiling belongs to
the weaker fresh-context channel. At either bound, any discrimination
there may be is not detectable at achievable power for untrained,
zero-shot elicitation.

\paragraph{The wider pool agrees.} In fresh context, none of the 28
sweep settings is significantly positive (Fisher exact), and the only
two that reach significance are \emph{negative}. Discrimination does rise with report
rate, which scopes the bound to the regime we observe. The in-episode
protocol tests that upper regime directly. Asking again inside the
episode lifts report rates to 78--99\% in four of its eight settings,
with Llama-70B going from 10 fresh reports to labeling 78\% of its
\Rp{} failures, and the odds ratio is still 1.06
(Appendix~\ref{app:detectdetail}).

\paragraph{Report style is a lineage trait, not a state readout.}
Across the fourteen-model English sweep, self-reported TOT
spans more than two orders of magnitude, from 4 for Gemma-3-27B to
1,468 for Mistral-7B, while the \Rp{} count stays within a factor of four (132--521). Among the
six models alone, self-reports run from 10 to 1,142 and \Rp{} counts
from 187 to 395 (Table~\ref{tab:sweep}). The style persists across
scale, consistent with post-training shaping verbalized confidence
\citep{leng2024taming,gandhi2025cognitive,kirichenko2025abstention,
han2025personality}. Concurrent work finds family-dependent
\emph{calibration} \citep{cacioli2026mmb}; we find a family-general
behavioral pattern beneath these reporting differences.

\section{Failure Is Computed, Not Said}
\label{sec:probe}
The behavioral pattern is robust but not reflected in verbal reports.
Is it legible anywhere else? We compare four signals derived from the
model on the same items: candidate entropy over the twenty sampled
answers \citep{farquhar2024detecting}; P(True), the model's
probability that its own answer is correct \citep{kadavath2022know};
a logistic probe on hidden states; and the model's own prospective
feeling of knowing (FOK). Each is scored by its area under the ROC
curve (AUC), where 0.5 is chance; the probe's is nested
(Appendix~\ref{app:detectdetail}).

The split is between what is computed
and what is said, not between what is internal and what is not. Every
signal derived from the model discriminates failed from successful
productions (0.52--0.84, every interval excluding 0.5), with candidate
entropy and P(True) above the hidden-state probe. The
\emph{verbalized} FOK, a rating from 0 to 100, sits at chance. Even a
full-range elicitation designed to spread it reaches only 0.50--0.59
against production and 0.48--0.56 against the recognition-defined
failure type (\Rp{} versus \Rn{}).

The channels are not timing-matched, so the
comparison needs care. The hidden-state probe and the token-level
FOK, the probability the model puts on the token \emph{Yes}, can be
read without generating an answer at all. The verbalized rating is
elicited before the naming attempt but needs a generated response of
its own. Candidate entropy needs the samples it is computed over, and
P(True) scores an answer the model has already produced. Both cost
more than the rest: entropy reads twenty draws, and P(True) needs an
answer first and then a pass to score it, while the probe, the
token-level FOK and the verbalized rating each take one pass. The
comparison matched on that count is the verbalized rating against the
two single-pass model signals, and it goes the same way. All three
are available before the target word is produced, but only the probe
and the token-level FOK are reliably above chance. The contrast is
therefore between computed and verbalized signals at the point each
becomes available, not between two readouts of one moment.

Two limits matter. Which \emph{kind} of failure an item is,
recognition-positive or recognition-negative, is not readable at these
sample sizes by any channel, with one partial exception in English.
And ``readable'' does not mean self-knowledge. A four-feature
item-difficulty baseline beats every nested Chinese probe. A
cross-model consensus predictor, the mean failure of the other five
models on the item, beats every probe. We therefore read the probe as a
deployment signal rather than as evidence of an internal state report
(Appendix~\ref{app:detectdetail}).

\section{Discussion and Conclusion}
\label{sec:discussion}
What survives the controls is availability without accessibility, and the strongest evidence is generative. Words twenty-one sampled attempts never surfaced are produced once a single form unit is
supplied, on 50 points more of the English failures and 30 more of the
Chinese ones than a matched retry that withholds the cue and nothing else. The cue is
not just pushing the model toward any word that begins
that way. Aimed at a pseudoword it returns almost nothing; aimed at a real word
that is not the answer it returns that word sometimes; aimed at the
target it returns the target 2.0--2.9 times as often in English and
1.6--2.0 in Chinese, both scored alike. A bare difficulty gap between generation and
recognition predicts none of it (Appendix~\ref{app:discdetail}). What this
supports is specific: knowledge sufficient to recognize the word under
adversarial options and to regenerate it from one unit of its form is
present while unconstrained naming fails, robustly in English and in Chinese to
the edge of detectability.

Two literatures meet here. Against the hidden-knowledge line
\citep{azaria2023internal,calderon2026empty,gekhman2026thinking},
these tasks add what a factual-QA design cannot: because the object
withheld is a single lexical form, availability can be supported by a causal within-item intervention rather than inferred from probes, and
the same design exposes the state--report dissociation. Against the
human literature, our availability is Tulving's
\citep{tulving1966availability}: cue-dependent retrievability over a broad class, not experience.

We asked whether a language model is ever in anything like the state a
speaker is in when a word will not come. Answering it needed a dataset
that did not exist, so we built and release one. The answer, behaviorally, is yes:
the word a model cannot name is often one unit of its own form away from
being named. What the model does not reliably do is say so --- the second
result, not a caveat:
a system that must know when a name is unsafe should measure the model,
not ask it.

\section*{Limitations}
\label{sec:limitations}
\paragraph{Power.} The English \Rn{} group is the binding constraint at
70B+ scale. Two enrichment sets aimed at it both failed for a diagnosable
reason: recognition is solved by definition-to-option matching, so rarer
targets draw \emph{more} eliminable distractors rather than fewer.
Larger \Rn{} groups need harder distractors, not rarer targets
(Appendix~\ref{app:taxdetail}).

\paragraph{Item validity.} Screening and reversal judging both use
Qwen2.5-72B, so the dataset inherits whatever that model gets wrong. Four
checks bound the exposure: a directional bias test, cross-family
re-judging, a cross-family rebuild of the answer key, and
re-certification of every paraphrase on the prompt text the models
actually saw, itself replicated with a second-family judge. None of
them inverts a ratio (Appendix~\ref{app:battery}).

There is a human floor as well: three graduate students independently
completed the 210-item audit blind to the judge except on the
reversal block, which reviews its verdict, and their
majority agrees with it on the three comparison-based instruments but
far less on uniqueness, where the machine certificates prove
systematically looser than calibrated human judgment
(Appendix~\ref{app:battery}). Answer-key uncertainty does reach the
headline interval, which widens without approaching parity
(\S\ref{sec:taxonomy}), but we propagate it for the \Rp{}:\Rn{} ratio only.

\paragraph{No human condition.} Every human comparison here is a literature
anchor drawn on different materials, populations and elicitation
protocols, and our difficulty-enriched items fall outside the
vocabulary any published TOT norm covers. Model--human gaps may
therefore reflect materials as much as models. A same-materials human
condition is the decisive missing control. It requires familiarity
norming of these items, in-episode elicitation and ethics approval, and
we leave it to future work.

\paragraph{Scope of the claim.} Claims cover instruction-tuned
models on single-word, definition-cued retrieval in two languages, and
the fourteen-model replication covers only the taxonomy and
report-style analyses. The 4AFC criterion is not by itself a lexical
diagnosis: run on non-lexical facts it labels arithmetic as
recognizable-but-unproduced. The cue-based availability test, however,
does not fire there (\S\ref{sec:construct}). We therefore read the
lexical result as cue-reachable, target-specific form information, and
the non-lexical generalization as the weaker claim its own cue condition
supports (Appendix~\ref{app:discdetail}). The pseudoword floor bounds
less than its size suggests: it conflates \emph{not stored}
with \emph{not a word}, and the decoy control measures the real-word case.

\paragraph{Composition versus retrieval.} The pseudoword and decoy comparisons bound what a cue can \emph{reach} rather
than separating retrieval from composition outright, because a
constraint-satisfaction account ranging over stored vocabulary
predicts the zero pseudoword floor as surely as a storage account
does. The decoy comparison is shared ground for the same reason: the decoy
is by construction the worse definition fit, so both accounts predict it.
What these comparisons do establish is that the cue reaches stored
real-word forms but not the tested pseudowords under this prompt, and that at matched definition fit it still prefers the stored target in
English. That preference narrows the constrained-search account rather
than excluding it: matching on fit leaves training co-occurrence,
frequency, tokenization and paraphrase-specific association free to
favor the target under either account, and the English estimate rests on
44 source items with a wide interval while the Chinese one is null.
Separating the two accounts would need a representation-level readout of
the target before the cue, which we do not have here.

\paragraph{Chinese is a boundary condition, not a replication.} Under
the adversarial and reversal filters imposed jointly, the English
dissociation holds while Chinese compresses to the edge of
detectability (\S\ref{sec:bounds}). One setting falls below parity with
an interval covering it either way, so we draw no directional
conclusion from that setting. The instrument gate that rescues it was
applied post hoc, and we report it on the pool as well as on the setting.
It also rests on an assumption we cannot test, that the reversal
judge's false-negative rate on failed items matches its rate on known
ones (Appendix~\ref{app:taxdetail}).

There is no frequency- or pool-matched cross-lingual comparison here
either. Every English--Chinese contrast is confounded with item pool,
frequency instrument and register, so our cross-linguistic differences
are differences between two \emph{datasets} as much as between two
languages.

\section*{Ethics Statement}
All stimulus data are dictionary entries and psycholinguistic norms.
The only human involvement was annotation by three of the
authors: a 210-item audit and a uniqueness check of English failure
items. From the annotators we collected task
judgments about dictionary items and no personal or sensitive
information. We caution against
interpreting ``TOT''
language as evidence of subjective experience; our claims are
operational throughout. Licensing and release
details are in Appendix~\ref{app:release}.

\section*{Acknowledgments}
This research utilised Queen Mary's Apocrita HPC facility, supported by
QMUL Research-IT. \url{http://doi.org/10.5281/zenodo.438045}

\bibliography{references}

\appendix
\section{The Dataset and Tasks in Full}
\label{app:battery}
This section gives the full item-construction, judge-annotation, stage, and statistical
detail summarized in \S\ref{sec:method}. In the appendices, as in the released records, the matched retry of
\S\ref{sec:construct} is the \emph{retest} and a wrong first unit is the
\emph{wrong-unit} condition (\emph{plac.} in Table~\ref{tab:cue}).

\paragraph{English items.}
From WordNet noun glosses we built two pools. The \emph{normed} pool
(8,226 candidates) requires SUBTLEX-US frequency \citep{brysbaert2009moving},
Kuperman AoA \citep{kuperman2012age}, concreteness
\citep{brysbaert2014concreteness}, and CMU pronunciations, with Zipf
1.0--4.5. A pilot showed modern LLMs produce \(\sim\)90\% of SUBTLEX-band
targets, so a \emph{beyond-corpus} pool (4,710 candidates) draws WordNet
lemmas absent from SUBTLEX, graded by \texttt{wordfreq} Zipf 1.0--3.2.
Quality filters remove proper nouns, taxonomic filler definitions, and
morphological leakage (any definition token sharing a long prefix with the
target, e.g.\ \emph{abstention}/\emph{abstaining}). The full-run set samples
501 normed + 1,002 beyond-corpus items, stratified by frequency.

\paragraph{Chinese items.}
The Chinese source is moedict (Taiwan MoE Revised Dictionary), from
which we built a SUBTLEX-CH-normed pool \citep{cai2010subtlex} (2,947
after proper-name filtering) and a beyond-corpus pool (4,000; absent
from SUBTLEX-CH). Definitions are segment-filtered to remove Latin script
and loanword metalanguage (\zh{音译} ``transliteration of\ldots''),
which otherwise leaks answers. The full-run set samples 402 + 501
items.

\emph{Conversion and its two documented confounds.} Titles and primary
definitions are converted Traditional\(\to\)Simplified with
\textbf{OpenCC} under the \texttt{t2s} configuration
(\texttt{opencc.OpenCC("t2s")} in the released
\texttt{build\_items\_zh\_hard.py}), applied independently to the
headword and to its definition. OpenCC \texttt{t2s} is many-to-one at the
character level: the traditional forms of \emph{to emit} and
\emph{hair} both map to \zh{发}, and three distinct traditional
characters map to \zh{干}. Distinct traditional headwords can
therefore collide on one simplified string, and the item builder keys
its definition map on the \emph{simplified} form, keeping the
first-occurring entry and silently discarding the later entry's definition.
Auditing the released builder bounds the loss. The first-wins rule
drops 642 of 143,702 candidate entries (0.45\%), and 5 of the 903 dataset
items (0.55\%) share a simplified key with a dropped competitor, so
for those five the retained definition may define a sense the writing system no longer distinguishes.
Separately, the frequency instrument
\citep{cai2010subtlex} is a \emph{mainland} norm applied to a
\emph{Taiwan}-lexicon item source. That register and lexical-inventory
mismatch is a documented confound in Chinese lexical research and a
live alternative explanation for part of the large En--Zh failure-rate
gap that \S\ref{sec:taxonomy} otherwise attributes to design. A
Taiwan-specific or register-marked target will receive a mainland
frequency estimate that understates its rarity for the models. We therefore read the Chinese ratio cautiously; this mismatch may be part
of why it is lower.

\paragraph{Judge annotation and contamination defense.}
Qwen2.5-72B annotates every item \citep{yi2026irony,lin2025augmenting}: (i)~synonym elicitation with
per-candidate verification (``could a dictionary use this definition
for word X?''), yielding vetted acceptable-answer sets (mean 2.8 EN /
1.9 ZH answers); (ii)~a uniqueness verdict (82\% EN / 59\% ZH unique);
and (iii)~a paraphrase of every definition that avoids the target and its
morphology. All production prompts use paraphrased definitions, so surface
forms match no indexed dictionary.

\emph{Certificate--prompt alignment.} Uniqueness certificates are
issued on the \textbf{paraphrase} the models actually see, not only
the original definition. Re-issuing on the prompt text agrees with the original-definition verdict on 91.5\% (En) / 88.6\% (Zh),
and a production rerun on original vs.\ paraphrased definitions
(En/Qwen-7B: 26.3\% vs.\ 29.4\% strict failure, 86.4\% item-level
agreement) bounds the label perturbation at 13.6\%. That
perturbation is propagated into the headline interval: the bootstrap
of \S\ref{sec:taxonomy} optionally re-draws each item's certificate at the two-judge error rate (10.0\%).
Re-drawn instead at the asymmetric rates the human audit measures
(46\%/62\% withdrawal of granted certificates, 0\%/25\% overturn of
refusals), the intervals move to [7.0,\,14.9]/[3.3,\,4.9], still nowhere near parity. On the
2,303 items whose paraphrase genuinely differs (1,502 En, 801 Zh), the
failure mode that threatens the dataset (unique on the original,
\emph{not} unique under the paraphrase the model saw) runs 4.2\% (63/1,502) and 7.2\% (58/801), against 65 and
33 in the harmless direction. Vetted answer sets are identical for
82.5\% of English items and identical-or-superset for
\(\approx\)91\%. Because the re-issuing judge wrote the paraphrases,
we repeated the check on a stratified 400-item English sample with
\emph{Llama-3.1-70B}. Verdicts agree on 90.0\%, and the two
judges' unique-rates differ by five points in the loosening direction
(Llama 88.7\%, Qwen 83.8\%), so the primary judge is the stricter of the
two. Dropping all 121 uniqueness-losing items and recomputing over the six leaves the headline essentially unchanged (English
\textbf{11.2:1\(\to\)12.1:1}, Chinese \textbf{4.9:1\(\to\)4.7:1}). The
two shifts run in \emph{opposite} directions, which is noise rather
than a shared instrument artifact.

Three further checks bound judge exposure. A directional bias test
argues against Qwen favoritism: if the judge favored Qwen-family knowledge, Qwen should fail \emph{less}, yet it fails more than the
size-matched Llama at both tiers (29.4 vs 23.3\%; 18.7 vs 14.5\%).
Cross-family re-judging by Llama-3.1-70B agrees at
\(\kappa{=}0.76\)--0.85 on the stratified samples and
\(\kappa{=}0.76\)--0.83 on the full-coverage run
(Appendix~\ref{app:taxdetail}). The 210-item human audit is
complete. Three graduate students judged every
item independently, blind to item-level judge verdicts except on the
reversal block, which shows the verdict and asks whether it is right, after a disclosed calibration pass
(a draft round exposed threshold misalignment on uniqueness, Fleiss
\(\kappa{=}0.13\); the criterion was rewritten as a fixed three-step
procedure and the 60 uniqueness and production items re-annotated
independently, \(\kappa{=}0.74\)). Inter-annotator agreement is Fleiss
\(\kappa{=}0.68\) overall. Each annotator matches the judge on 78\%--91\%
of items, and the majority verdict matches it on 86\%: 92\% on synonym
acceptability, 96\% on paraphrase quality, 92\% on reversal review, 80\%
on the ten production spot checks, but 66\% on uniqueness. On a roughly
balanced subsample the majority refuses 14 of 26 granted certificates
(En 6/13, Zh 8/13) and overturns 3 of 24 refusals, a liberal bias
consistent with treating certificates as perturbable rather than gold.

\emph{Annotator instructions.} Each audit item asked one yes/no
question: whether the definition alone is clear enough to single out
its target (uniqueness); whether a candidate synonym is an acceptable
answer to the definition; whether a paraphrase keeps the meaning, does
not leak the target and is similarly difficult; whether the judge's
verdict on a reversal item is right; and whether a production answer
scored wrong should have been accepted as a synonym or variant. The
three-step uniqueness procedure stops at the first step that decides:
(1)~no, if the definition gives only a broad category and a generic
feature; (2)~no, if the annotator can name another common word, other
than a spelling variant or derivative of the target, that fits every
component of the definition exactly; (3)~otherwise yes. A definition
that identifies its target by an encyclopedic fact rather than by its
meaning also gets no.

\emph{Rebuilding the answer keys cross-family.} Llama-3.1-70B
re-derived the acceptable-answer sets for a stratified 299-item
English sample. The keys are identical for 61\% of items, and the
Llama key is a superset for a further 28\%. Re-scoring production
against the Llama key flips 5.4--7.4\% of strict-failure labels, and
the modal sampled answer of an originally-failed item is accepted for
9--28\%, highest for the Llama family under a Llama-built key (the
expected family-phrasing effect). The decisive statistic is
state-conditional: originally-\Rp{} items cease to be strict failures at
17--38\% and originally-\Rn{} items at 33--43\%. That English difference
is \emph{not detectable} rather than shown comparable (\Rn{}-minus-\Rp{}
\(+26.0/+4.5/+14.0/-4.5\) points, all Fisher \(p\geq0.12\), pooled odds ratio (OR) 1.66 [0.35,\,7.89]). With 5--7 \Rn{} items per model the test cannot
distinguish ``comparable'' from ``twice as high''. Under \emph{full}
relabeling, production is re-scored against the Llama key and the
classifier re-applied, so items that \emph{become} strict failures are
admitted rather than only those that leave; the sample ratios are then
13.2/11.0/9.8/5.8:1, none inverting. These are ratios on the 299-item
\emph{sample} and not comparable to Table~\ref{tab:taxonomy}'s full-dataset
ratios, so we claim nothing about which settings rise. The Chinese
rebuild (stratified 300 items) moves keys more, as expected where
judge agreement is lower: keys are identical for 29\% of items,
strict-failure labels flip on 17--26\%, and modal answers are accepted
for 4.5--17.7\%. Chinese is where ``comparable'' is earned. With only 27--46 \Rn{} items, the state-conditional flip rates are
12.1--35.4\% (\Rp{}) vs.\ 10.9--35.0\% (\Rn{}), pooled OR 0.76
[0.42,\,1.39], and the difference, such as it is, runs the protective
way. Fully relabeled sample ratios are 5.1/2.9/5.7/3.2:1, none
inverted. All keys are released.

\emph{Propagating the rebuilt key to item-level analyses.} The
aggregate-ratio defense does not by itself protect analyses that
condition on item-level labels, so we re-ran the cue-rescue state
contrast on the rebuild samples under the fully relabeled key. It
survives and \emph{strengthens}: pooled Chinese logit \(b{=}+0.668\rightarrow+1.204\) [\(+0.475,+1.934\)], \(p{=}0.0012\),
\(n{=}581\), with three of four Chinese models individually
significant and English positive in 4/4 (\(+14.3\) to \(+40.2\)
points), though underpowered at \Rn{} \(n{=}3\)--4.

\paragraph{Materials inventory and annotation coverage.}
The release contains the \textbf{main dataset} of 2,406 items (1,503 En =
501 normed + 1,002 beyond-corpus; 903 Zh = 402 + 501); a 500-item
English ultrarare extension; a 500-item opaque-word set; 799
pseudoword controls; a 300-item English non-lexical set; and
option-set overlays for the adversarial-distractor, decoy-target
(borrowed-filler and option-matched), decoy-cue and pseudoword-cue
conditions, with raw per-setting records and the analysis scripts. 

\paragraph{Conjunctive availability criterion.} ``Availability'' is \emph{operationalized} by recognition alone in the main taxonomy
(what supports the interpretation is the three-test analysis of
\S\ref{sec:construct}--\ref{sec:bounds}), but
the reversal control offers a stricter conjunctive test, counting an item as \Rp{} only if it is a strict production failure
\emph{and} passes recognition \emph{and} the
word\(\rightarrow\)definition reversal check. The reversal control discriminates strongly and in the
right direction in every setting (\Rp{}-minus-\Rn{} pass rate \(+17.6\) to
\(+43.8\) points, all Fisher \(p\leq0.0019\)). Under the conjunctive
rule the ratios fall to En 10.5/10.8/5.2/6.3:1 and Zh
3.5/3.7/1.2/1.8:1 (model order Qwen2.5-7B / Qwen2.5-72B / Llama-3.1-8B
/ Llama-3.1-70B in both sets). That costs 19--61\% of the \Rp{} group,
and the worst case (Zh Llama-3.1-8B, also the lowest recognition
ceiling) collapses to 1.2:1, but \emph{no ratio inverts under this
criterion applied on its own} (composed with the adversarial
condition, one Chinese setting does on the comparable denominator and
three on the literal one; Appendix~\ref{app:taxdetail}). S1--S6 themselves run on all six models. The extra analyses that run on
four are the judged reversal, a second pass of S5 asked inside the
failure episode and its logit readouts, P(True), the strong-elicitation
FOK, the circumlocution pair, and the target-absent, decoy, pseudoword
and hidden-state probe stages: 8 of the 12 settings each.

\paragraph{S1 Production.} Given the (paraphrased) definition, name the word:
greedy, plus \(n{=}20\) samples at \(T{=}1.0\). \emph{Strict failure} = greedy wrong and 0/20, scored against the item's
vetted answer set throughout, so no acceptable answer appeared at all; the 1--2/20 and
3--5/20 bands are excluded from binary contrasts and released with the per-item records.

\paragraph{Decoding sensitivity.}
Production samples at \(T{=}1.0\) and top-\(p\) 1.0, a broad,
untruncated regime chosen so that low-probability targets are not
excluded. Strict failure depends on decoding, so we reran production
for Qwen2.5-7B and Llama-3.1-8B in both languages at \(T{=}0.7\) and
at top-\(p\) 0.9, with the same greedy decode plus 20 samples.
Recognition is item-level and is reused. Tighter decoding raises the
strict-failure rate by 3.4--4.3 points in English and by at most 2.3
points in Chinese, and keeps 98.1--99.7\% of the original failure set
(Table~\ref{tab:decoding}). The \Rp{}:\Rn{} ratio remains similar in
every setting, and on the failures common to both regimes the cue and
retest rates change by less than one point. The sample budget can be
varied without new runs, by truncating the 20 stored samples. Pooled
over the six models, as the budget grows from the greedy decode plus 1
sample to plus 20, the English strict-failure rate falls from 35.1\% to
22.3\% and the \Rp{}:\Rn{} ratio from 14.8 to 11.2, flattening after ten
samples (12.1, 11.5, 11.2 at 10, 15, 20). In Chinese the rate falls
from 75.5\% to 63.6\% and the ratio from 5.4 to 4.9.

\begin{table}[t]
\centering
\scriptsize
\setlength{\tabcolsep}{2.9pt}
\begin{tabular}{lllrrrrr}
\toprule
Lang & Model & Decoding & Fail\% & Kept & \Rp{}:\Rn{} & Cue & Retest \\
\midrule
EN & Qwen2.5-7B   & \(T{=}1.0\) & 29.4 & --      & 13.2 & 53.9 & 11.3 \\
   &              & \(T{=}0.7\) & 33.7 & 440/442 & 14.7 & 54.1 & 11.1 \\
   &              & top-\(p\) 0.9 & 32.9 & 439/442 & 13.4 & 54.0 & 11.1 \\
EN & Llama-3.1-8B & \(T{=}1.0\) & 23.3 & --      &  7.1 & 64.0 & 13.5 \\
   &              & \(T{=}0.7\) & 27.4 & 346/350 &  7.7 & 63.5 & 12.8 \\
   &              & top-\(p\) 0.9 & 26.7 & 346/350 &  6.8 & 63.5 & 13.1 \\
\midrule
ZH & Qwen2.5-7B   & \(T{=}1.0\) & 71.7 & --      &  5.4 & 18.8 &  4.4 \\
   &              & \(T{=}0.7\) & 74.0 & 645/647 &  5.4 & 18.6 &  4.1 \\
   &              & top-\(p\) 0.9 & 73.6 & 641/647 &  5.6 & 18.8 &  4.0 \\
ZH & Llama-3.1-8B & \(T{=}1.0\) & 69.1 & --      &  3.1 & 30.0 &  5.3 \\
   &              & \(T{=}0.7\) & 70.7 & 612/624 &  3.2 & 29.5 &  4.9 \\
   &              & top-\(p\) 0.9 & 69.1 & 613/624 &  3.2 & 29.3 &  4.5 \\
\bottomrule
\end{tabular}
\caption{Production rerun under tighter decoding. \(T{=}1.0\) is the
original regime (top-\(p\) 1.0). \emph{Fail\%} = strict production
failures (greedy wrong and 0/20) as a percentage of the dataset (1,503
En / 903 Zh). \emph{Kept} = original strict failures that remain
strict failures under the rerun. \Rp{}:\Rn{} uses the original
recognition records. \emph{Cue} / \emph{Retest} = any-of-20 rates (\%)
from the original cue and retest records, over the \Rp{} and \Rn{}
failures common to the original run and the rerun (all of them for
\(T{=}1.0\)).}
\label{tab:decoding}
\end{table}

\paragraph{S2 FOK (all items, prospective).} The probe is asked
without any retrieval attempt or mention of failure: ``If you were
shown four candidate words, how confident are you that you could pick
the exact word this definition describes?'' FOK is measured as (a) a
verbalized 0--100 rating, (b) the token-level probability of
\emph{Yes} in a yes/no variant, and (c) post hoc, the self-consistency
of the S1 samples.

\paragraph{S3 Recognition.} 4-alternative forced choice; distractors
are experimenter-built (an orthographic
neighbor, a definition-overlap semantic associate, a frequency- and
length-matched random word), never model-generated
\citep{west2023paradox}. Distractors come from the item's own source pool and exclude the target
and, in English, its WordNet synonyms. The orthographic neighbor
minimizes Levenshtein distance to the target, over letters in English
and characters in Chinese, among words at most three edits away and
within two units of its length; ties go to the closer log frequency.
The semantic associate maximizes the number of content tokens its
definition shares with the target's (English: non-stopwords of three or
more letters; Chinese: characters other than 16 function characters),
with at least one required. The random word is drawn with a fixed seed
from words within 0.35 log\(_{10}\) units of the target's frequency and
three units of its length. A slot with no qualifying candidate takes a
random pool word, as the neighbor slot does in 362 of 1,488 English
items. An option set is dropped when the judge later accepts one of its
distractors as an answer (15 English and 10 Chinese items, which have no
recognition record). The adversarial set
(Appendix~\ref{app:taxdetail}) takes, from both pools, the three words
with the highest IDF-weighted overlap with the displayed definition,
excluding answers, their morphological variants, and words one edit
from an answer. The target appears once in each
position (Latin square); each rotation is scored by the first-token
probability over A--D. Success = \(\geq\)3/4; \(\leq\)1/4 = failure; 2/4 excluded. The four rotations are not independent draws: on 96.1\% of English and
93.2\% of Chinese strict failures the model picks the same word in at
least three of them. The chance level that respects this keeps each
item's four choices and permutes which option is the target; pooled
over the six models it passes 24.0\% (En) and 23.3\% (Zh) of strict
failures, against 88.7\% and 78.4\% observed. The scoring mode is innocuous: in those eight settings, every greedy generation
that parses to a letter matches the first-token argmax (38,071/38,071
rotations). Across all recognition runs in the project, including
supplementary models and unparseable generations, agreement is
99.8\%.

\paragraph{S4 In-episode partial access.} A methodological finding forced a
protocol revision. Asked partial-form questions in fresh context, models
answer about \emph{whichever candidate the definition brings to mind} instead of
the target. A model that produces \emph{adventure} correctly will still
guess ``E'' as its first letter because it is thinking of \emph{escapade},
and \zh{下厨} elicits \zh{烹} (from \zh{烹饪}). So we probe \emph{inside
the production episode}. After the model's wrong answer, the experimenter
replies ``\emph{X} is not the word I am looking for. Even if you cannot
retrieve it, what is your best guess for its first letter [length /
syllable count / a word it sounds like]?'' The negative feedback is true,
and it pins the referent exactly as in \citet{brown1966tip}, where probes
occur during the TOT episode. On items the model \emph{can} produce, the
same protocol yields first-letter accuracy 100\% (vs.\ 34\%--46\% fresh-context). \emph{A confound in that validation,
stated plainly:} in the known-item condition the model has just emitted
the correct word, so the probe is close to a copying task and 1.00 is
expected whether or not the protocol ``pins the referent''. On \Rp{} items,
by contrast, the referent is pinned only by truthful negative feedback
about a \emph{wrong} answer. The ceiling therefore shows the protocol works in the easy case,
while the fresh-context comparison diagnoses referent drift rather
than certifying the probe on \Rp{} items. The stage enters this paper
as the instrument that pins the referent for the
in-episode self-report of \S\ref{sec:mimicry}. 

\paragraph{S5 Self-report.} Run on all items: ``answer; or say TOT
if you know it but cannot retrieve it; or DK''. It never contributes
to state classification.

\paragraph{S6 Reversal control.} The model defines each target word
(word\(\rightarrow\)definition). Qwen2.5-72B judges correctness against
the definition.

\paragraph{Full statistical specification.} Analyses are logistic
GLMs with standardized covariates (Zipf, length, and subword token
counts computed per model) and setting fixed effects. Pooled analyses cluster standard errors (SEs) by item, and per-model fits are released with
the code.

\paragraph{Reproducibility.} All inference uses vLLM 0.15.0
\citep{kwon2023vllm}. Single-shot stages decode greedily; production
sampling uses \(T{=}1.0\), \(n{=}20\), top-\(p\) 1.0. Recognition uses a
fixed \(4\times4\) Latin square. Model IDs, token caps, judge and
distractor procedures, prompts, judgments, per-item generations for every
setting, and seeded analysis code ship with the release. One episode is released with its full stage records: item
\emph{taps}
(Llama-3.1-70B, En), greedy \emph{lightsout} 0/21, recognition 4/4, correct cue resolving production 20/20 against a 0/20
wrong-unit condition, and no TOT report.

\section{Prompts}
\label{app:prompts}
We print every English prompt verbatim below, followed by the Chinese
self-report templates, which are load-bearing for
\S\ref{sec:mimicry}; the remaining Chinese prompts are direct
translations and ship with the code. Production (system): \emph{``You are a helpful
assistant. Answer with a single English word only, no punctuation or
explanation.''} (user): \emph{``What is the English word that this definition
describes?\;/\;Definition: ``\{gloss\}''\;/\;Answer with the single word
only.''} FOK (verbalized): \emph{``Consider this definition: ``\{gloss\}''.
It describes a specific English word. If you were shown four candidate words,
how confident are you that you could pick the exact word this definition
describes? Reply with a single integer from 0 (certain I could not) to 100
(certain I could). Reply with the number only.''} In-episode partial access
(after the model's wrong answer $w$): \emph{```$w$' is not the word I am
looking for. I mean one specific English word. Even if you cannot retrieve
it, what is your best guess for its FIRST LETTER? Reply with a single letter
only.''} (length, syllable, and sounds-like variants analogous). Recognition: \emph{``Consider this definition:
``\{gloss\}''\;/\;Which of the following words does it describe? A.\ \ldots\ D.\
\ldots\ Answer with the letter only.''} Self-report: \emph{``Consider this definition:
``\{gloss\}''\;/\;Try to name the specific English word it
describes. If you can retrieve the word, reply with
the word only. If you feel you know the word but cannot retrieve it right
now, reply exactly `TOT'. If you simply do not know the word, reply exactly
`DK'.''} Reversal control: \emph{``What does the
English word ``\{word\}'' mean? Answer with one short definition sentence
only.''} Cue: \emph{``\ldots Hint: the word begins with the letter
`\{c\}'.''} The matched \emph{retest} condition has no prompt of its
own: it is the production prompt above, re-run on the same items
with the sampling seed offset, so that its twenty draws are
independent of the twenty that defined the failure rather than a
replay of them. Chain-of-thought retry (the condition the retest replaces,
same turn structure as the cue condition and scored against the same
answer key, with no form information):
\emph{``What is the English word that this definition describes?\;/\;Definition: ``\{gloss\}''\;/\;Think step by step: first
reason about the meaning, then list a few candidate words, then end
your reply with exactly one line of the form ANSWER: <the single
word>.''}

\paragraph{Chinese self-report templates (verbatim), and FOK.}
Chinese self-report volume is a lineage trait. Across the sweep it ranges from 0/903 to 899/903, so there
is no uniform silence. Because the near-zero rates of Qwen2.5-72B (23/903), Llama-3.1-8B
(19/903) and Llama-3.1-70B (0/903) are a headline result, the exact Chinese wording is
printed here rather than deferred to the code, so that the silence cannot
be attributed to the prompt. \textbf{Fresh-context self-report}
(\texttt{SELFREPORT\_ZH}):
\emph{``\zh{看这个释义：“\{gloss\}”}\;/\;\zh{请说出它描述的那个中文词语。}\;/\;\zh{- 如果你能想起这个词，只回答这个词。}\;/\;\zh{- 如果你感觉自己认识这个词但此刻想不起来，只回答「TOT」。}\;/\;\zh{- 如果你根本不知道这个词，只回答「DK」。}''}
\textbf{In-episode self-report} (\texttt{SELFREPORT\_EP\_Q\_ZH}, asked
after the model's rejected wrong answer):
\emph{``\zh{先把你的猜测放在一边：现在，对于我所指的那个词，你的状态是哪一种？}\;/\;\zh{- 如果你现在能想起这个词，只回答这个词。}\;/\;\zh{- 如果你感觉自己认识这个词但此刻想不起来，只回答「TOT」。}\;/\;\zh{- 如果你根本不知道这个词，只回答「DK」。}''}
\textbf{FOK, yes/no} (\texttt{FOK\_YESNO\_ZH}, the source of P(yes))
translates the English \emph{``\ldots If you were shown four candidate
words, would you recognize the exact word this definition describes? Answer Yes or
No only.''} \texttt{FOK\_USER\_ZH} translates the verbalized English prompt
above. Both ship with the code.
\textbf{Label rendering.} The self-report labels are the Latin strings
\texttt{TOT} and \texttt{DK} in both languages, never translated, so the
scored token is identical across languages. Their definitions are given in
Chinese in every variant. We cannot rule out that a Latin-script label in
a Chinese prompt depresses the Chinese report rate, but the in-episode
re-elicitation (\S\ref{sec:mimicry}) uses the same rendering and raises
Chinese report rates to 84--99\% in three of four settings, which argues that rendering is not the
binding factor.

\section{Taxonomy Details}
\label{app:taxdetail}
\begin{table}[!htb]
\centering
\scriptsize
\setlength{\tabcolsep}{2.6pt}
\begin{tabular}{llrrrrr}
\toprule
Lang & Model & Fail\% & \Rp{} & \Rn{} & Ceil\% & OR \\
\midrule
EN & Qwen2.5-7B & 29.4 & 395 & 30 & 97.8 & 1.03 \\
EN & Qwen2.5-14B & 24.7 & 331 & 26 & 97.9 & 1.89 \\
EN & Qwen2.5-32B & 23.3 & 321 & 20 & 98.8 & 1.28 \\
EN & Qwen2.5-72B & 18.7 & 253 & 18 & 98.5 & 0.59 \\
EN & Llama-3.1-8B & 23.3 & 292 & 41 & 97.9 & 0.58 \\
EN & Llama-3.1-70B & 14.5 & 187 & 24 & 98.2 & 0.07* \\
EN & Gemma-3-12B & 33.2 & 445 & 37 & 96.8 & 0.64 \\
EN & Gemma-3-27B & 26.8 & 364 & 29 & 97.0 & 0.08 \\
EN & GLM-4-9B & 27.9 & 328 & 53 & 93.7 & 0.59 \\
EN & Mistral-7B & 39.0 & 521 & 47 & 94.9 & 0.72 \\
EN & OLMo-2-13B & 23.6 & 305 & 31 & 95.6 & 1.21 \\
EN & OLMo-2-32B & 19.9 & 264 & 26 & 98.4 & 0.99 \\
EN & Phi-4 & 18.4 & 247 & 20 & 97.9 & 1.63 \\
EN & R1-Distill-14B & 11.6 & 132 & 27 & 91.1 & 0.78 \\
\midrule
ZH & Qwen2.5-7B & 71.7 & 513 & 95 & 94.8 & 0.86 \\
ZH & Qwen2.5-14B & 64.5 & 469 & 80 & 94.6 & 0.71 \\
ZH & Qwen2.5-32B & 61.0 & 469 & 63 & 95.1 & 1.26 \\
ZH & Qwen2.5-72B & 57.4 & 423 & 75 & 93.3 & 1.12 \\
ZH & Llama-3.1-8B & 69.1 & 429 & 137 & 87.2 & 0.74 \\
ZH & Llama-3.1-70B & 58.0 & 386 & 101 & 91.8 & n/a$^\dagger$ \\
ZH & Gemma-3-12B & 72.1 & 486 & 126 & 89.4 & 1.02 \\
ZH & Gemma-3-27B & 69.4 & 487 & 85 & 91.5 & n/a$^\dagger$ \\
ZH & GLM-4-9B & 64.5 & 442 & 92 & 90.7 & 0.52* \\
ZH & Mistral-7B & 86.9 & 346 & 297 & 69.8 & 0.50 \\
ZH & OLMo-2-13B & 78.8 & 285 & 199 & 74.7 & 1.44 \\
ZH & OLMo-2-32B & 78.3 & 496 & 143 & 89.1 & 1.33 \\
ZH & Phi-4 & 67.6 & 462 & 113 & 91.1 & 0.57 \\
ZH & R1-Distill-14B & 55.6 & 391 & 75 & 91.8 & n/a$^\dagger$ \\
\bottomrule
\end{tabular}
\caption{The fourteen-model sweep: fourteen models from seven lineages
$\times$ two languages, 28 settings. \emph{Fail\%} = strict
production failure; \emph{\Rp{}}/\emph{\Rn{}} = strict failures passing or
failing the scored 4AFC ($\geq$3/4 and $\leq$1/4 rotations), in
items; \emph{Ceil\%} = known-item recognition accuracy, the instrument
check, gated at 85\%; \emph{OR} = the base-rate-free odds ratio for a
self-reported TOT label discriminating \Rp{} from \Rn{} among
strict failures. $^{*}$ Fisher $p<0.05$; $^{\dagger}$ no self-reports
among classified strict failures, so undefined. Report counts and the
excluded 2/4 band ship with the release.}
\label{tab:sweep}
\end{table}

\paragraph{The sweep.} Table~\ref{tab:sweep} runs the taxonomy over
fourteen models from seven lineages. \Rp{} exceeds \Rn{} in all 28
settings, including the two that fail the 85\% instrument gate
(Mistral-7B Zh at 69.8\% and OLMo-2-13B Zh at 74.7\%, reported but
not counted), so the gate protects the ratio's magnitude rather than
its direction. No odds ratio is significantly positive, and the two
that reach significance are negative (\S\ref{sec:mimicry}). Within
the six models of Table~\ref{tab:taxonomy}, failure falls as models
grow while the ratio moves non-monotonically, so size buys production
without shifting the composition of what remains in any one
direction.

\paragraph{\Rn{} scarcity, and a failed remedy.} \Rn{} failures are intrinsically
scarce, and two enrichment sets failed to change that. A 500-item
beyond-corpus English extension added 643 \Rp{} but only 23 \Rn{} across the
four models. A second set of 500 morphologically opaque rare words
(\emph{dittany}, \emph{tenesmus}, \emph{muntjac}; median Zipf 1.56
against 2.23) made it worse, yielding 635 \Rp{} and 18 \Rn{}. The instrument
is sound throughout (known-item ceilings 98.1--100\%), so the mechanism
is diagnosable. Recognition is solved by definition-to-option semantic
matching, not by form familiarity, and rarer targets draw distractors
from more distant domains that can be eliminated without knowing the
word. Each ``harder'' set therefore pushes recognition further to
ceiling (84.4\(\to\)92.2\(\to\)93.5\% of strict failures scoring 4/4).
A 1.67\(\times\) larger English item pool bought only 41 further \Rn{}
observations (113 to 154). The binding factor is \emph{distractor
confusability} rather than target rarity, so larger English \Rn{} groups need within-domain co-hyponym distractors. Rebuilding the extension's
4AFC with adversarial distractors confirms this from the other side,
compressing its ratio from 23--46:1 to 4--8:1 without inverting it.

\begin{figure}[t]
\centering
\includegraphics[width=\columnwidth]{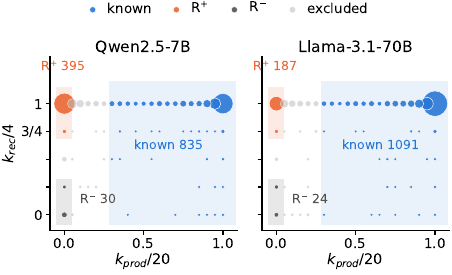}
\caption{Behavioral space, English (\(n{=}1{,}488\); 15 further items
have no recognition record). Production over S1's 20 samples on x, 4AFC
recognition over S3's four rotations on y. Both axes are lattices, so
one marker is drawn per occupied position with area proportional to the
items there; the largest holds 568. Shading marks the label regions,
each annotated with its total. \emph{known} is
\(k_{\text{prod}}\geq6/20\). \Rp{} and \Rn{} additionally require a
strict production failure, which is \(k_{\text{prod}}{=}0\)
\emph{and} a wrong greedy answer, so the left-hand column is not by
itself the rule; \Rp{} then needs \(k_{\text{rec}}\geq3/4\) and \Rn{}
\(k_{\text{rec}}\leq1/4\). The rest is excluded.}
\label{fig:space}
\end{figure}
\paragraph{The headline ratio by item pool.} If recognition were solved
by definition-to-option elimination alone, the beyond-corpus pool should
inflate the ratio, since its targets draw distractors from more distant
domains. Stratified by the normed / beyond-corpus pool indicator, it
does in English: normed 8.5:1 [4.9,\,17.9] (390/46) against
beyond-corpus 12.3:1 [8.8,\,17.9] (1,389/113), intervals that overlap
almost entirely. In Chinese it \emph{reverses}, with normed 7.3:1
[5.2,\,11.8] (783/108) against beyond-corpus 4.3:1 [3.6,\,5.2]
(1,906/443). \Rp{} outnumbers \Rn{} at least fourfold in all four strata and
P(\Rp{}\,\(\mid\)\,classified failure) stays within 81\%--93\%, so the
dissociation is not an artifact of the beyond-corpus majority. The
English contrast is too imprecise to exclude a modest inflation there.

\paragraph{Adversarial distractors.} To test whether the \Rp{}:\Rn{} ratio
merely reflects easy distractors, we rebuilt every 4AFC with \emph{all
three} distractors drawn from the highest-definition-overlap semantic neighbors
in the pools (IDF-weighted, answer-set-excluded) and re-ran recognition.
Under this adversarial regime the known-item ceiling stays at 89.8--98.1\%
(the sets remain fair). The ratio compresses, En pooled 11.2:1\(\to\)5.7:1
[4.6,\,7.2] and Zh 4.9:1\(\to\)3.2:1 [2.8,\,3.7] across all six
models, and the dissociation remains large everywhere. The ratio's magnitude moves with the operationalization
(\S\ref{sec:bounds}). The asymmetry itself does
not.

\begin{table}[t]
\centering
\scriptsize
\setlength{\tabcolsep}{3.2pt}
\begin{tabular}{llrrrr}
\toprule
& & \multicolumn{2}{c}{Pseudoword (A)} & \multicolumn{2}{c}{Real-word decoy} \\
\cmidrule(lr){3-4}\cmidrule(lr){5-6}
Lang & Model & std.\ (\%) & adv.\ (\%) & borrowed (B) & matched (B$'$) \\
\midrule
En & Qwen2.5-7B    & 29.8 & 27.2 & 52.7 & \textbf{32.2} \\
En & Llama-3.1-8B  &  7.2 &  4.2 & 70.8 & \textbf{39.7} \\
En & Qwen2.5-72B   & 38.8 & 24.8 & 72.5 & \textbf{41.2} \\
En & Llama-3.1-70B & 12.0 & 10.8 & 51.0 & \textbf{29.0} \\
\midrule
Zh & Qwen2.5-7B    & 28.1 & 16.8 & 49.2 & \textbf{30.0} \\
Zh & Llama-3.1-8B  & 21.1 & 10.5 & 52.0 & \textbf{27.6} \\
Zh & Qwen2.5-72B   & 38.1 & 20.8 & 61.0 & \textbf{35.7} \\
Zh & Llama-3.1-70B & 42.9 & 26.1 & 50.5 & \textbf{30.0} \\
\bottomrule
\end{tabular}
\caption{Target-absent controls, on the four models that carry them: how often a
model consistently endorses an option ($\geq$3/4 rotations) when the correct answer has been removed. All four
columns are percentages. \textbf{A} replaces the target with a novel
form, so it bounds the \emph{novel-form} elimination channel from
below (std.\ = scored options, adv.\ = adversarial options); being
silent about familiar words, it cannot bound definition-inference in the
direction a reader wants. \textbf{B} and \textbf{B$'$} replace the
target with the item's strongest definition-compatible non-answer, so they
bound the \emph{real-word} channel from above; B uses borrowed
fillers, while B$'$ keeps each item's own three scored distractors,
making its option structure identical to S3's, which is why B$'$ is
the bound we carry. Per-setting counts and matched-population cue rates
are in the release.}
\label{tab:controls}
\end{table}

\paragraph{Target-absent controls (A, B, B\('\)).}
The scored 4AFC could in principle be passed by definition-compatibility
inference rather than stored knowledge, so we measured how often each
model endorses an option when the correct answer is \emph{absent}. Three
variants differ only in what replaces the target and how the remaining
options are built. Control~\textbf{A} substitutes a novel form
(letter-trigram pseudowords / novel two-character compounds built from
\emph{other} items' words). Controls~\textbf{B} and~\textbf{B}\('\)
substitute the item's strongest definition-compatible non-answer, with B\('\)
keeping each item's own three scored distractors so that its option
structure is identical to S3's. Table~\ref{tab:controls} carries all eight
model\(\times\)language settings of each and the direction each control
bounds. Three properties of the table matter for what comes later. B exceeds
B\('\) in 8/8 settings, by 19 to 31 points, so option structure and not
inference alone drives the difference, and B\('\) is the bound we
carry. B\('\) does not fall at scale (27.6--39.7\% at 7--8B against
29.0--41.2\% at 70B/72B), which means the margin argument is not a
small-model artifact. And family ordering reverses between tiers:
real-word inference and novel-form elimination are not one channel, and
we read no family generalization into either. Control~A additionally
shows that even the highest floor sits far below real recognition (En
88--94\%, Zh 76--88\%), and that part of Chinese ``recognition'' is
compositional, one reason we treat the Chinese ratio as the
conservative bound. Substituting B\('\) for A in the floor-correction
estimator, which credits every decoy pick as a spurious \Rp{}, gives
6.0:1 (En) and 2.7:1 (Zh) against 10.0:1 and 4.3:1 uncorrected. The
cue tests sit well below these floors: the same first-unit cue reaches
a decoy on 14--21\% (En) and 11--29\% (Zh) of items, and a pseudoword
on 0/400 (En) and 0/399 (Zh) in 15 of 16 conditions.

\paragraph{Inference control C: what the criterion rests
on.} A trial that deletes the target measures how well a model guesses in
the target's absence, so controls A and B cannot \emph{establish} the
criterion. What they supply is a same-format measurement of the
inference-only baseline, and control~C is the \emph{margin} comparison
against the same option structure with the target restored.
Control~B\('\) puts that baseline at 27.6\%--41.2\%, against 87.7\%--93.4\% (En)
and 75.8\%--84.9\% (Zh) with the target restored to the same option
structure: a per-setting margin of 48--61 points, unchanged by the four
settings added at 70B/72B. Option structure is matched. One mismatch
remains, and we name it rather than assert equivalence: the populations
differ (a 400 En / 400 Zh subset yielding 345 / 333 usable records,
against the full dataset restricted to strict failures). The margin is an
\emph{upper} bound for the separate reason that the decoy is a judged
\emph{non}-answer. A definition-inferring model is also \emph{not} expected
to lose the adversarial condition, since the target \emph{is} the
best-fitting option, so that condition answers a different question,
distractor easiness, with a compression bound of 5.7:1 (En) / 3.2:1 (Zh)
and no inversion. The test needing no options at all is the one we lean
on hardest (\S\ref{sec:construct}, Test~3).

\paragraph{The two target-present criteria imposed jointly.} \Rpsub{joint} = strict
production failure \(\wedge\) adversarial recognition
(\(k_{\text{rec}}\geq3/4\) against the maximally confusable option
set) \(\wedge\) a correct word\(\rightarrow\)definition reversal, scored
against two denominators, one comparable to Test~1 and one deliberately
punitive (Table~\ref{tab:joint}). English survives on the comparable
denominator with room to spare (pooled 3.90:1 against the matched baseline of 10.0:1; every setting \(\geq\)2.84:1) and stays
above 1.2:1 even on the punitive literal denominator. Chinese compresses
much harder: the pool falls to 1.66:1 comparable and to 0.65:1 on the
literal denominator, with Zh Llama-3.1-8B below parity on both as
printed (0.88 / 0.35). That setting's bootstrap interval covers parity in
every reading we have (0.88 [0.70,\,1.06] comparable, 1.12 [0.89,\,1.39]
corrected), so nothing in the paper turns on which reading is right. The
pooled item-cluster intervals are 1.66:1 [1.42,\,1.96] for the Chinese
comparable ratio and, on the literal denominator, 1.70:1
[1.42,\,2.03] (En) and 0.65:1 [0.58,\,0.73] (Zh). The
two filters are close to multiplicative rather than redundant. The
binding one is the reversal control, not the distractors: of the 429
standard Chinese Llama-8B \Rp{} items, 178 pass adversarial recognition but
fail reversal against 25 the other way round, while the known-item
adversarial ceiling stays at 89.8--95.2\% (En 95.9--98.1\%). The
collapse is not a distractor artifact.

\emph{Not a judge-severity artifact.} The natural defense is that the
Qwen2.5-72B reversal judge under-credits Chinese definitions. It is
disconfirmed. A cross-family Llama-3.1-70B judge (full coverage for
Qwen2.5-7B and Llama-3.1-8B in both languages) moves reversal pass rates
\emph{up} in English (\(+4.3\) pp) and slightly \emph{down} in Chinese
(\(-2.2\) pp), widening the very gap the artifact hypothesis says it
should narrow, at \(\kappa{=}0.76\)--0.83. Every Chinese setting that
inverted still inverts (Zh Llama-8B 0.88\(\to\)0.85), and even crediting
an item if \emph{either} judge passes it leaves it inverted (0.96
comparable, 0.40 literal). Two gaps remain: cross-judge verdicts do not
cover the 72B/70B models, and reversal judgments exist only for the four models that carry
them.

\emph{Reversal accuracy by state.} Judged
word\(\rightarrow\)definition accuracy shows the expected gradient
(known 78--96\%, \Rp{} 39--81\%, \Rn{} 15--47\%) and does predict state
(logit \(b{=}+1.15\), \(p<10^{-5}\)). Adding the item's reversal
correctness as a covariate leaves the cue effect intact (logit
\(b{=}+4.27\); \S\ref{sec:bounds}).

\emph{The reversal instrument needs the same ceiling gate we impose on
recognition, and one setting fails it.} We gate recognition at an 85\%
known-item ceiling but applied the reversal filter (the binding one
here) without an analogous gate. Applying one post hoc uses each model's
reversal pass rate on words it produces correctly. All four English settings clear it (91.5\%--92.7\%). Chinese does not: Qwen-7B (92.6\%) and Qwen-72B
(95.7\%) pass, Llama-3.1-70B is 87.6\% [82.4\%,\,91.5\%] straddling the gate,
and \textbf{Llama-3.1-8B is 78.0\% [70.0\%,\,84.3\%], failing outright} (the
cross-family judge harsher still, 73.2\%). A filter that rejects 22\% of
items the model demonstrably knows removes availability-positive items
for reasons orthogonal to availability, so by our own rule that setting is
instrument-invalid. Correcting each setting for its own known-item
false-negative rate moves English conjunctive ratios 10.5/10.8/5.2/6.3
\(\to\) 11.5/11.7/5.7/6.9, Chinese 3.5/3.7/1.2/1.8 \(\to\)
3.7/3.9/1.6/2.1, and the joint comparable column 2.05/2.43/0.88/1.67
\(\to\) 2.22/2.54/1.12/1.91: the two Chinese Qwen settings barely move and
\textbf{the single inverted setting crosses back above parity}
(0.88\(\to\)1.12). The correction assumes what it cannot test, that the
judge's false-negative rate is the same on words the model cannot
produce as on words it can, and our own numbers question that (En
p\(_{\text{known}}\) 91.5\%--92.7\% against p\(_{\text{R}^{+}}\) 72.9\%--81.3\%).
So the correction is not load-bearing: the setting it rescues covers parity
either way.

The defensible conclusion is that, \emph{jointly imposed, the two
target-present criteria compress the ratio hard (English to 3.9:1
[3.15,\,5.02], Chinese to between 1.66 and 2.05:1 depending on how the
gate is applied, both on the comparable denominator) and invert on no
setting whose interval excludes parity}.
The gate is applied to the pool, not only to the setting it rescues, so we
print every reading: the Chinese pool is 1.66:1 uncorrected,
\textbf{1.85:1} re-pooling the four corrected settings, \textbf{2.05:1}
excluding the gate-failing setting outright. English moves the same way,
3.90 to 4.25:1 with no setting excluded. Every pooled reading on that
denominator is above parity. The
literal \Rnsub{joint} denominator still falls below parity in
three Chinese settings and we continue to report it as the punitive
accounting it is. Read strictly, the Chinese conclusion rests on three
instrument-valid settings rather than four.

\paragraph{Control D: floor-corrected and excluded-band ratio bounds.}
Two further sensitivities bound the headline ratios. (a) Treating each model's pseudoword-4AFC floor \(f\) as the rate at which true-\Rn{} items pass
the scored 4AFC without stored knowledge (observed
\Rn{}\,=\,true-\Rn{}\,\((1{-}f)\); spurious \Rp{}\,=\,true-\Rn{}\,\(\cdot f\)) and
solving per model: corrected ratios are 6.5--9.0 (En, pooled 7.6:1) and
1.8--3.6 (Zh, pooled 2.6:1), compressed but never inverted. (b)
Allocating the excluded 2/4-recognition band entirely to \Rn{} (the conservative extreme; 74 En / 206 Zh items) gives 7.6:1 (En) and 3.6:1 (Zh).
Under every operationalization \emph{applied singly} (standard,
adversarial, floor-corrected, band-conservative, conjunctive),
recognition-positive failures outnumber recognition-negative ones
several-fold in every pool (per setting the conjunctive rule bottoms out at
1.2:1, Zh Llama-3.1-8B). The composition that compresses hardest is the
joint criterion in Chinese, where the pool falls to 1.66--2.05:1
depending on how the reversal gate is applied; on the punitive literal
denominator three Chinese settings remain below parity
(Table~\ref{tab:joint}). Correction~(a) is scoped by control~A: it removes
the novel-form elimination channel and nothing else. It does not bound
the real-word inference channel measured in control~B. That channel is
addressed by the discriminative tests of control~C, and the strictest
floor we can measure, substituting B\('\)'s rate, gives the 6.0:1 /
2.7:1 bound above.

\paragraph{Cross-lineage replication.} OLMo-2-13B
\citep{olmo2025two,groeneveld2024olmo}, a third, lineage-independent
family with fully open training data, replicates the taxonomy in
English (failure 23.6\%, \Rp{}:\Rn{} \(\approx\)10:1), addressing both
family-specificity and training-data-transparency concerns at once;
its Chinese setting fails the recognition-ceiling instrument check
and is not counted. In total, fourteen models from seven lineages
(adding Mistral-7B, Gemma-3-12B/27B, GLM-4-9B, Phi-4, OLMo-2-32B, and
R1-Distill-14B, a Qwen2.5-14B distillation counted in the Qwen
lineage) replicate the taxonomy, with English strict-failure spanning
11.6--39.0\% and \Rp{} outnumbering \Rn{} everywhere (Table~\ref{tab:sweep}).
One instrument check gates this claim. Recognition must lower-bound
knowledge, so we count a setting as a replication only when its
known-item ceiling exceeds 85\%. Mistral-7B Zh (69.8\%) and
OLMo-2-13B Zh (74.7\%) fail the check; their settings are reported but
not counted. The tally is identical at an 80\% gate and survives a
90\% one (23/28 counted, all fourteen models still represented).
Labels are also stable across rotation pairs: split-half agreement
runs 0.94--0.96 in English and 0.83--0.94 in Chinese. The
R1-Distill-14B/Qwen2.5-14B pair is a controlled same-base contrast,
in which reasoning-focused post-training halves the failure rate
(11.6 vs 24.7\%) while also changing report style, evidence that both
the capability and the metacognitive voice are shaped by
post-training.

\section{Cue Details}
\label{app:sigdetail}
\begin{table}[t]
\centering
\scriptsize
\setlength{\tabcolsep}{2.1pt}
\begin{tabular}{llrrrrrrrr}
\toprule
& & \multicolumn{4}{c}{\Rp{}} & \multicolumn{4}{c}{\Rn{}} \\
Lang & Model & \(n\) & retest & cue & plac. & \(n\) & retest & cue & plac. \\
\midrule
EN & Qwen-7B & 395 & 12.2 & 55.2 & 4.3 & 30 & 0.0 & 40.0 & 3.3 \\
EN & Qwen-14B & 331 & 8.5 & 64.4 & 3.3 & 26 & 3.8 & 42.3 & 0.0 \\
EN & Qwen-32B & 321 & 12.1 & 63.9 & 6.9 & 20 & 0.0 & 50.0 & 0.0 \\
EN & Qwen-72B & 253 & 10.7 & 64.4 & 4.7 & 18 & 5.6 & 55.6 & 0.0 \\
EN & Llama-8B & 292 & 14.0 & 65.1 & 6.5 & 41 & 9.8 & 56.1 & 2.4 \\
EN & Llama-70B & 187 & 15.5 & 68.4 & 5.3 & 24 & 0.0 & 50.0 & 0.0 \\
\midrule
ZH & Qwen-7B & 513 & 4.7 & 20.3 & 0.6 & 95 & 3.2 & 10.5 & 1.1 \\
ZH & Qwen-14B & 469 & 3.2 & 39.7 & 1.9 & 80 & 2.5 & 36.2 & 2.5 \\
ZH & Qwen-32B & 469 & 3.6 & 39.9 & 1.9 & 63 & 1.6 & 28.6 & 4.8 \\
ZH & Qwen-72B & 423 & 4.3 & 44.2 & 0.9 & 75 & 2.7 & 29.3 & 1.3 \\
ZH & Llama-8B & 429 & 6.1 & 35.2 & 1.2 & 137 & 2.9 & 16.1 & 0.7 \\
ZH & Llama-70B & 386 & 8.3 & 42.7 & 1.0 & 101 & 1.0 & 34.7 & 0.0 \\
\midrule
EN & pooled & 1779 & 11.9 & \textbf{62.8} & 5.1 & 159 & 3.8 & \textbf{49.1} & 1.3 \\
ZH & pooled & 2689 & 4.9 & \textbf{36.4} & 1.3 & 551 & 2.4 & \textbf{24.7} & 1.5 \\
\bottomrule
\end{tabular}
\caption{Cue-unlocked production (Test~3), over all twelve
model\(\times\)language settings. The conditions run on one matched population; \(n\) is the items in
that state and all rates are percentages of it. \emph{retest} = the same production prompt re-run on the same items
with an independent seed, which withholds the cue and nothing else;
\emph{cue} = production after the correct first letter or character, in the greedy decode or any of the 20 samples;
\emph{plac.} = after a wrong first unit. Uncued production is
\(0/n\) in every setting by construction. The chain-of-thought retry,
which the retest replaces, is reported in Appendix~\ref{app:sigdetail}.
Pooled rates, the increment, and item-cluster intervals by state are in
Table~\ref{tab:validity} and Appendix~\ref{app:sigdetail}.}
\label{tab:cue}
\end{table}
\paragraph{Cue-unlocked production, per setting.} Pooled cue-condition
rates with item-cluster bootstrap intervals are 62.8\%
[58.9,\,66.4] for English \Rp{} and 49.1\% [35.6,\,63.1] for English
\Rn{}, 36.4\% [33.5,\,39.4] for Chinese \Rp{} and 24.7\%
[19.5,\,30.6] for Chinese \Rn{}. Test~3 of
\S\ref{sec:construct} is reported here at full resolution so that it
can be re-scored like every other test. The data are six models \(\times\) two languages, 5{,}178 setting\(\times\)item rows,
every row a strict production failure whose pre-cue production is
verified at \(k_{\text{prod}}{=}0/20\) with greedy wrong. ``Unlocked'' = the target
appears in the greedy decode or in any of the 20 samples after the cue turn; the wrong-unit cue supplies a wrong first unit.

Table~\ref{tab:cue} gives every setting: the correct first unit
resolves a majority of English \Rp{} items and a large minority of
Chinese ones, the retest far fewer, and the wrong-unit cue almost
none. \emph{The counterfactual is a
fresh attempt, not the wrong-unit condition and not the zero.} The retest condition
reprompts the identical items with the identical prompt and an
independent seed, and resolves 11.2\% (En) / 4.5\% (Zh) of all
cleanly classified failures. That rate is the size of the
regression-to-the-mean term, which the selection on
\(k_{\text{prod}}{=}0/20\) makes unavoidable and which no condition
measured against the zero itself could have shown. The correct first
unit's contribution is the increment over that condition. It is taken on
the 20 samples of each arm, since the retest's greedy decode only replays
the original failure: 61.6\% against 11.2\% (En) and 34.3\% against 4.5\%
(Zh), that is \(+50.3\) points
[\(+46.7,+53.8\)] (En) and \(+29.8\) [\(+27.2,+32.6\)] (Zh),
individually positive with the interval excluding zero in 12/12
settings (item-block bootstrap; smallest \(+14.3\), largest
\(+54.6\) points).

An earlier version of this comparison used a no-cue chain-of-thought
retry, which recovers 34.8\% (En) / 7.3\% (Zh). That condition confounds
three things with the cue at once: it changes the instruction, it
changes the output format, and it draws 5 samples where the cue condition
draws 20. The retest removes all three. Pricing chain-of-thought on its own then
means putting both conditions on one budget, which the retest records make
exact: keeping a random 5 of its 20 draws is hypergeometric, so no
independence assumption is needed. At that budget the retest sits at
3.8\% (En) / 1.5\% (Zh) and chain-of-thought is worth \(+31.1\)
points [\(+28.2,+34.1\)] in English and \(+5.8\)
[\(+4.8,+6.9\)] in Chinese. Nearly all of what the chain-of-thought baseline recovered in English was therefore chain-of-thought rather than a
re-draw; in Chinese there was little of either.

\emph{The increment on items verified as uniquely answerable.}
Scoring against an acceptable-answer set removes the case where the
model produced a different but equally good word, since such an item
is not a failure at all. What it cannot remove is a competitor the
judge itself missed, for which the cue would be disambiguating rather
than retrieving. The cross-family key rebuild gives a direct test:
over its stratified 299-item English sample, Qwen2.5-72B returns 245
uniqueness verdicts and Llama-3.1-70B 266, agreeing on 237. On the
matched pair restricted to those items the retest recovers 15.3\% and
the correct first unit 68.6\%, an increment of \(+53.3\) points
[\(+44.4,+62.0\)] over 84 source items and 229 setting\(\times\)item
rows, against \(+50.3\) [\(+46.7,+53.8\)] on the full English
population. The retest baseline rises too, as unambiguous items should
be easier to redraw, and the cue rises further, so the gap widens
rather than closing. Three annotators then judged all 647 English
strict-failure items by hand, under the audit's three-step uniqueness
procedure and blind to model outputs. Pairwise agreement is 75--90\%
(Fleiss \(\kappa{=}0.74\)); a majority confirms 330 of the 640 classified
items as uniquely answerable and rejects 310. On the confirmed items
the increment is \(+50.5\) points [\(+45.7,+55.6\)] over
924 setting\(\times\)item rows. It is \(+51.2\) [\(+46.0,+56.5\)] on the
277 items confirmed unanimously and \(+50.1\) [\(+44.9,+55.1\)] on the 310
rejected ones, and the \Rp{}:\Rn{} ratio on the confirmed items is
9.9:1 (839 to 85 rows) against 11.2:1 overall. The rebuild covers English only, so there is no
Chinese counterpart.

\emph{And the wrong-unit cue suppresses rather than resets:} a wrong first
unit drives production to 4.8\% (En) / 1.3\% (Zh), below the matched
retest by \(-6.4\) points [\(-8.3,-4.7\)] (En) and \(-3.2\)
[\(-4.1,-2.3\)] (Zh), so a false form constraint forbids the target.
Measured against the chain-of-thought condition the same suppression reads
\(-17.2\) points, but that comparison charges the wrong-unit condition for the chain-of-thought the baseline enjoyed and the wrong-unit condition
did not. The published cue-specificity log-odds is \(+3.54\)
[\(+3.31,+3.78\)] over those four models and \(+3.49\) [\(+3.30,+3.67\)]
over all twelve (\(I^2{=}0\)--1.9\%). It is therefore a
correct-versus-\emph{suppressed} contrast, and it overstates what the
correct cue supplies by exactly the amount the wrong unit suppresses. We report it as a specificity statistic (the cue must be the
\emph{right} unit) and the cue-minus-retest increment as the unlocking
statistic.

\emph{State grading, and one fit for state, condition and interaction.}
The state contrast is taken on the correct-cue condition, the
wrong-unit cue being nearly eventless. The state, condition and
interaction estimates below come from one
joint model: \texttt{resolved} \(\sim\) state \(+\) condition \(+\)
state\(\times\)condition \(+\) z(Zipf) \(+\) z(length) \(+\) z(tokens)
\(+\) setting fixed effects, item-clustered SEs, one row per (item, condition) for the correct-cue and wrong-unit conditions
(the no-cue retest enters the paired increment tests, not this fit), all twelve settings, 10,356 condition-rows over 1,363 items,
reference level \Rn{} on the wrong-unit condition. The fit gives a wrong-unit condition
state contrast of \(+0.602\) (SE 0.486) [\(-0.352,+1.555\)],
\(p{=}0.216\); a cue main effect of \(+3.689\) [\(+2.766,+4.611\)];
state\(\times\)cue \(+0.005\) (SE 0.480) [\(-0.936,+0.947\)],
\(p{=}0.992\); and a correct-condition state contrast of \(+0.607\) (SE
0.150) [\(+0.312,+0.901\)], \(p{=}10^{-4}\), which reconciles with
the released per-condition pools (\(+0.619\) fixed-effect, \(+0.553\)
random-effects). The two-stage restatement gives \(+0.55\)
[\(+0.28,+0.85\)] under the item-block bootstrap, 1.6\(\times\)
wider than an inverse-variance interval that would treat twelve
settings over the \emph{same} items as independent studies
(\(\bar\rho{=}0.15\)); the corrected fit shows no residual heterogeneity (\(Q{=}10.48\), \(p{=}0.49\)) and per-setting log-odds are positive
in 12/12 (clustered \(p{=}1.5\times10^{-4}\)). Narrower model sets move
the interaction but never its conclusion, \(-0.228\)
[\(-1.288,+0.831\)] on both languages together and
\(-0.630\) [\(-2.068,+0.807\)] English-only, wide and covering zero either way. The interaction null is not evidence for storage
over composition, since a constraint-satisfaction solver over
definition-plus-cue predicts a state-independent cue effect just as
a graded-availability account does. The condition that separates them is the subject of the next paragraph.

\paragraph{The decoy-cue condition: the test's target-absent analogue.}
Test~3 needs its own target-absent analogue. On the decoy items of
control~B\('\) (Appendix~\ref{app:taxdetail}) we ran the identical
forced-cue stage, supplying the decoy's first letter or character
(the decoy is a real word, not the stored answer for that definition)
plus the same wrong-unit condition. Decoy production, in the greedy decode or any of the 20 samples, runs at \textbf{15.1\%} / \textbf{13.2\%} /
\textbf{16.9\%} / \textbf{18.1\%} across En Qwen-7B / Llama-8B /
Qwen-72B / Llama-70B and at \textbf{10.4\%} / \textbf{14.6\%} /
\textbf{29.2\%} / \textbf{27.7\%} in Chinese, with greedy at
4.1\%--9.1\% (En) and 4.1\%--20.0\% (Zh). The wrong-unit cue sits at \textbf{0\%--1\%} in all eight settings, so the suppression
replicates on decoys and is not an artifact of the target's own
availability.

Comparing a decoy rate over all usable decoy items against a
real-target rate over clean failures would flatter us, so both sides
are recomputed on one population: decoy items whose source item is a
clean failure, with the real-target rate taken on exactly those items. Matching
moves the decoy rate by at most 3.9 points in either direction
(matched decoy 14.1\%/16.3\%/20.8\%/18.3\% En, 10.5\%/15.7\%/28.6\%/25.2\% Zh;
real-target 53.1\%/68.6\%/65.3\%/73.3\% En, 20.3\%/33.3\%/46.2\%/42.5\% Zh), leaving
margins of 3.1--4.2\(\times\) (En) and 1.6--2.1\(\times\) (Zh).
Those margins are not scoring-symmetric, because the target is
credited against the vetted answer set and the decoy on a single
string. We therefore restrict both sides to source items whose
answer set is a single word, making target credit exact-string as
well: \textbf{2.0--2.9\(\times\)} (En, pooled 2.46\(\times\), 113 rows) and \textbf{1.6--2.0\(\times\)} (Zh, pooled 1.78\(\times\), 850 rows), the numbers we carry. At Zh 72B the decoy reaches
28.6\% against 46.2\%, so in Chinese this condition bounds composition only loosely. The English matched
settings are small (60--128 items), the 400-item overlay sitting on a
dataset with a low strict-failure rate. Definition fit itself predicts decoy
unlock (logit \(b{=}+0.84\) En / \(+0.45\) Zh), so the composition
channel is real. At matched answer-level fit (\(\geq0.8\)) the
English target--decoy margin persists, 61.8\% vs.\ 32.9\% on 76 paired
rows over 44 source items, difference +28.9 points (95\% CI [+8.4,\,+48.1]), ratio
1.9\(\times\) [1.2,\,3.4] (source-item cluster bootstrap), while the
Chinese one vanishes (32.0\% vs.\ 30.8\%, +1.2 points [\(-12.5\),\,+14.5]),
which is one reason the strong reading is scoped to English.

\emph{Cohort size.} If the cue worked only by narrowing the search, its benefit would grow
with the number of candidates the first letter removes. An English
target's cohort is the number of CLEARPOND types
\citep{marian2012clearpond,brysbaert2009moving} sharing its first letter, entered as
\(\log_{10}(\text{count}+1)\). Over the 1,938 classified English failures
(640 items, six models), a
logistic regression of cue unlock on standardized log cohort (model
fixed effects, item-clustered SEs) gives \(b{=}-0.10\) per SD
[\(-0.27,+0.06\)]. Unlock is 65.3\%, 62.9\% and 56.1\% from the
smallest to the largest cohort tercile (medians 916, 1,553 and
2,766). Across the full range, 4 to 3,398 words, the fitted slope
implies \(-23.2\) points [\(-60.9,+14.4\)]. In the fit-matched
target-versus-decoy comparison (fit \(\geq0.8\)), adding log cohort
changes the target coefficient by 1.4\% (\(+1.21\) to \(+1.20\)).
Restricting those rows to the 14 first letters cued in
both arms, the target unlocks on 63.5\% of 63 rows and the decoy on 32.0\% of 75,
\(+31.5\) points [\(+10.1,+51.7\)] (source-item cluster
bootstrap). The restriction breaks the source-item pairing, so these
rows are unpaired. In Chinese (CC-CEDICT headwords sharing the first
character) the slope is \(b{=}+0.15\)
[\(+0.02,+0.27\)].

\emph{The floor: the pseudoword-cue condition.} A decoy is a real word, so it bounds composition of a \emph{stored}
form; the pseudoword condition probes novel strings with no intended
lexical entry.
Under the identical forced-cue stage on the pseudoword dataset (same
definition format, same cue format, first unit of a novel pseudoword or
novel two-character compound), the pseudoword is produced
\textbf{0/400} (En) and \textbf{0/399} (Zh) in fifteen of sixteen
conditions (four models \(\times\) two languages \(\times\) correct-cue and
wrong-unit condition), under greedy and any-of-20 decoding alike; English is 0/400
in all eight of its conditions. The sixteenth, Zh Llama-3.1-70B under the correct cue,
produced exactly one pseudoword across 399 items
(\textbf{1/399}\,=\,0.3\%, any-of-20 only; greedy 0/399), which we
print rather than round away. On the matched population the three
conditions are therefore pseudoword \(0\%\)--\(0.3\%\) \(\to\) decoy
\(11\%\)--\(29\%\) \(\to\) stored target \(20\%\)--\(73\%\) (pooled
Wilson 95\% intervals: pseudoword upper bounds 0.2\%/0.4\%, decoy 16.8\%
[13.2\%,\,21.1\%] / 19.3\% [16.9\%,\,21.9\%], target 63.0\% [57.8\%,\,67.9\%] / 34.6\%
[31.6\%,\,37.7\%] En/Zh; per-setting values in the release). The decoy step is shared ground; it is by construction the worse definition
fit, so both accounts predict it. The zero pseudoword floor is inconsistent with unrestricted string
assembly under this prompt, but not with constrained search over stored
lexical forms. The fit-matched target--decoy contrast is therefore the
discriminating test, and it supports the stronger interpretation in
English only.

Four scope limits. The real-word case is the decoy condition's to carry,
and where scale moves the decoy rate it moves it \emph{up} (Zh
10.4\%/14.6\% at 7--8B against 29.2\%/27.7\% at 70B/72B), which is why the
Chinese margin is thin. The floor itself is thin there too, the
Chinese pseudoword being a novel compound of \emph{stored}
characters where the English one shares no stored substring. The
decoy is a judged \emph{non}-answer and fits the definition less well,
making that condition a conservative rather than exact floor. And the released cue
records store outcome flags rather than raw text, so cue
\emph{resolution} is scored under the original answer key even where
the state label is rebuilt (Appendix~\ref{app:battery}); a broader key
would only credit more resolutions.

\section{Detection Details}
\label{app:detectdetail}
\paragraph{Signal-by-signal comparison.} The ranges over the eight settings with probe data are: candidate entropy 0.73--0.81 on failure and
0.46--0.66 on state; P(True) 0.72--0.84 and 0.47--0.61; the nested
hidden-state probe 0.58--0.72 and, on state, one of eight intervals
clearing 0.5; token-level FOK P(yes) 0.52--0.70, the one channel
with a partial state signal in English (three of eight intervals);
and the verbalized rating, which even under the strong elicitation
below reaches only 0.50--0.59 and 0.48--0.56. The item-difficulty
baseline reaches 0.60--0.61 (En) / 0.73--0.75 (Zh) and the
cross-model consensus predictor 0.86--0.93.

\paragraph{Self-consistency baselines.} The self-consistency of the
20 production samples (share agreeing with the modal candidate)
predicts greedy failure at an AUC of 0.70--0.80 across the twelve settings (six models \(\times\) two languages), indistinguishable
from candidate entropy over the same twelve (0.71--0.81; 0.73--0.81 above is those eight settings), whose near-monotone
transform it is. On \Rp{}-vs-\Rn{} it is similarly weak (0.46--0.66). A
one-bit variant (greedy answer agrees with the modal sample) is much
weaker (0.53--0.68). Both corroborate the division of labor in
\S\ref{sec:probe}: sampling-based signals excel at failure detection
but not at state discrimination.

\paragraph{P(True) and strong-elicitation verbalized FOK.}
Kadavath-style P(True), the token-level probability that the model's
own greedy answer is correct, is a strong failure detector (AUC 0.72--0.84 across those eight settings), on par with candidate
entropy and above the hidden-state probe, and like both state-blind
(\Rp{}-vs-\Rn{} 0.47--0.61, no CI excluding 0.5). A \emph{strong} verbalized
elicitation (0--100 probability with an explicit full-range,
anti-round-number instruction) does not rescue the channel:
failure AUC 0.50--0.59, state AUC 0.48--0.56, with responses still
collapsing onto 3--19 distinct values against 440--1,330 for P(True).
The verbalized channel's information loss is therefore not an
artifact of elicitation style.

\paragraph{Probe baselines and selectivity.}
We run the three controls the probing literature demands
\citep{hewitt2019designing}. (i) An item-feature logistic
baseline (Zipf, length, tokens, pool) predicts failure at AUC
0.60--0.61 (En) and 0.73--0.75 (Zh). (ii) A cross-model consensus
baseline, the mean failure of the other five models on the item, reaches 0.86--0.93, \emph{above} the hidden-state probe
(0.58--0.72 nested); most of what the probe reads is shared item
difficulty, linearly present in the prompt encoding. (iii) A
permuted-label selectivity control returns \(\approx\)0.51. We
therefore scope the probe claim: as a \emph{deployment} signal it
stands (one forward pass, no access to other models); as evidence of
\emph{self}-knowledge beyond shared difficulty it does not.

\paragraph{In-episode self-report.}
The referent-drift finding argues fresh-context self-report is
elicited on the wrong episode, so we re-ran S5 \emph{in-episode}
(after the rejected greedy attempt, exactly as S4). In-episode
elicitation transforms report \emph{rates}: Llama-3.1-70B En labels
78\% of its \Rp{} failures in-episode against 10 reports in 1,503
items fresh, and the three remaining Chinese settings label 84\%,
99\%, and 99\%, while Llama-3.1-8B Zh stays silent on every one. It does
not transform discrimination, with per-setting ORs 0.32--3.89, all
n.s., and a pooled OR of 1.06 [0.75,\,1.50], on 1,240 English and
2,159 Chinese classified failures. Across the wider
28-setting \emph{fresh-context} pool the two settings with a
significant association are the \emph{negative} ones (Llama-70B En
0.07, GLM-4-9B Zh 0.52). The odds ratio on the live subset of
\S\ref{sec:mimicry}, 0.90 [0.74,\,1.09], is an item-cluster
Mantel--Haenszel estimate. Three settings contain no self-reports among
classified strict failures and carry no discrimination information,
leaving 25 informative ones. Over these the random-effects estimate
is 0.856 [0.708,\,1.034] with \(I^2{=}19.4\%\). Discrimination rises
with report rate (log-odds slope \(+0.77\) [\(+0.09,+1.44\)]), and the
fitted OR at a 90\% report rate
is 1.23 [0.93,\,1.75], the top of the observed regime
(\S\ref{sec:mimicry}).

\paragraph{Nested-CV probe evaluation.} Layer-swept AUCs select the
read-out layer on the folds they report; nested evaluation (inner 3-fold
layer selection on outer training folds only, outer 5-fold pooled AUC,
1{,}000-resample item bootstrap CIs) gives, per setting (failure probe /
state probe): En Qwen-7B 0.63 [0.60,0.65] / 0.53 [0.42,0.63]; En Llama-8B 0.58
[0.55,0.61] / 0.62 [0.52,0.71]; En Qwen-72B 0.70 [0.67,0.72] / 0.52 [0.38,0.65]; En
Llama-70B 0.72 [0.69,0.75] / 0.58 [0.45,0.71]; Zh Qwen-7B 0.67 [0.62,0.72] / 0.51
[0.45,0.58]; Zh Llama-8B 0.65 [0.60,0.69] / 0.55 [0.50,0.61]; Zh Qwen-72B 0.67
[0.63,0.71] / 0.49 [0.41,0.56]; Zh Llama-70B 0.67 [0.62,0.71] / 0.51 [0.45,0.58].
Every failure-probe CI excludes 0.5; exactly one of eight state-probe CIs
does (En Llama-8B), which is about what eight tests produce by chance.
At 18--41 English \Rn{} items the state-probe question is underpowered,
not settled negative.

Under the earlier layer-swept protocol (read-out layer selected on
the folds it reports), failure probes reach 0.60--0.72 and
\emph{state} probes appear to reach 0.54--0.71. The nested evaluation
above shows the state signal does not survive honest layer selection
at current \Rn{} group sizes, so we claim no reliable decoding of
failure type. Failure probes peak in late layers (22--30 of 28--32 at 7--8B, 76--80 of 80 at 70B/72B),
consistent with retrieval commitment late in the stack
\citep{yeom2026commitment}, and transfer across languages within a
model (train-En\(\to\)test-Zh AUC 0.54--0.60 and conversely). Cost
sits on an access/latency axis: entropy needs a 20-sample budget per
item, the probe a single white-box forward pass.

\section{Discussion Details}
\label{app:discdetail}
\paragraph{Isn't this just ``generation is harder than recognition''?}
Partly, and deliberately so. The taxonomy \emph{starts} from that
asymmetry, as the human paradigm does; the content of the paper is
what distinguishes the availability state from a generic failure. A bare
difficulty gap predicts none of the following, all observed: matched,
experimenter-built distractors with a validated ceiling;
form-specific causal cue unlocking that a wrong unit suppresses
rather than reproduces, from a production floor of zero, and that a
decoy target reproduces only weakly and a pseudoword essentially not
at all (0\%--0.3\% \(\to\) 11\%--29\% \(\to\) 20\%--73\%, one matched
population, from pseudoword to decoy to target; \S\ref{sec:construct}); and a
self-report channel that fails to track any of it. The construct
label ``availability without accessibility'' is a claim about this
cluster of properties, not about difficulty or experience (\S\ref{sec:construct}).

\paragraph{Lemma or lexeme?} Two-stage models separate a
syntactic--semantic \emph{lemma} from its phonological \emph{lexeme}
\citep{levelt1999theory,dell1986spreading}, and human TOT is read as
lemma selected, lexeme incompletely activated \citep{burke1991tip}.
Our data show part of the shape a transmission deficit predicts
(definitional access intact while unconstrained naming fails) but no
evidence for the architecture. We never manipulate the lemma properties that
separate the levels in humans \citep{miozzo1997gender}, and a
subword-tokenized transformer has no level at which form is
retrieved as a unit once meaning is fixed.

\paragraph{Non-lexical control set: the asymmetry is not a format
artifact.} A sharper form of the same worry is that
recognition\(\gg\)production is generic to any 4AFC-versus-generation
comparison, whatever is being retrieved. The identical pipeline ran
on 300 English non-lexical facts (120 world capitals, 90 two-digit
products, 90 chemical elements) in all four of those models.
Strict-failure counts are nearly identical across the four (96--101
of 300); the \Rp{}:\Rn{} split (the 2/4 band counted as \Rn{} here) is not. Qwen2.5-7B splits 87/12 (7.3:1),
Qwen2.5-72B 100/1 (100:1), and Llama-3.1-70B 86/10 (8.6:1), but Llama-3.1-8B
splits 30 \Rp{} / 67 \Rn{}, 0.45:1, \Rn{}-dominant and \emph{inverted},
while the same model runs 7.1:1 \Rp{}-dominant on English lexical
items (Table~\ref{tab:taxonomy}; per-setting counts in the release). That single inversion
does the work. Holding format, pipeline, prompts and model fixed and
changing only the domain flips the direction of the asymmetry, so
the format cannot by itself manufacture \Rp{} dominance.
Format-\emph{sufficiency} is ruled out; format-\emph{relevance} is
not. The inversion is also \emph{Llama-3.1-8B-specific}. At 70B/72B
both families are \Rp{}-dominant on non-lexical facts, more sharply
than on the lexicon. And outside the lexicon we control neither
norms nor distractor construction nor cue form, so we read the other
three settings as consistent with a graded availability account rather
than as evidence for one. In three of four settings the criterion also
assigns availability labels where there is no lexical entry to be
available (two-digit multiplication).

We therefore ran the cue condition here too, and it is the sharper result.
The identical forced-cue stage unlocks production on \textbf{2.0\%}
of non-lexical \Rp{} items against a \textbf{1.3\%} wrong-unit condition
(6/303 vs.\ 4/303 pooled; per model 0\%--3.4\% correct against
0\%--2\% wrong-unit condition), where the same intervention on the English \Rp{} items reaches 63\% against 5\%. So the 4AFC criterion is not by itself
a lexical diagnosis, but the cue-based availability test is.
The cue test does not fire on material a model can re-derive from the
prompt, exactly as it should not, since a product is computable and
a word form is not. We read the lexical result as
availability without accessibility and the non-lexical generalization
as the
weaker claim its own cue condition supports, and we do not merge them.

\paragraph{Scope.}
Our claims cover instruction-tuned models retrieving single words
(overwhelmingly nouns) from dictionary-style definitions in two languages, across two model families (twelve settings), with the
taxonomy replicated across fourteen models from seven lineages. The state definition, thresholds and model set were fixed before
data collection (\S\ref{sec:stats}); we did not pre-register an
analysis plan. Naturalistic dialogue,
multi-word expressions, proper names, and verbs are open.
Portability splits in two. \emph{The taxonomy ports on generation
alone}. Production, recognition, the in-episode probes, and the cue
intervention need only text in and out, and the argmax-agreement
check (38,071/38,071 rotations) shows generated-letter scoring
changes nothing. \emph{The readout results do not}. FOK P(yes),
P(True), and first-token recognition scoring need token-level
logprobs most commercial APIs do not expose, and the hidden-state
probe needs white-box access. Instruction tuning is a standing
confound for the metacognitive results, since the missing report
could be a monitoring deficit or a tuning-induced style. Two
observations bound the confound: a failure signal exists in hidden states yet never reaches the report, and report styles differ
sharply between families with similar behavioral states. The
confound is not resolved without a base-model replication. Finally,
minimal new \Rn{} items under ultrarare enrichment and substantial cue-rescue
even of \Rn{} items both fit a graded-accessibility continuum on which
our operational states are regions rather than kinds
\citep{brown1991tot}; these tasks measure position on it without
committing to discreteness.

\section{Licensing and Release Details}
\label{app:release}
Code and data are available at \repourl.
Everything in the release that we author (item annotations, answer keys, judgments, prompts,
analysis and reconstruction code) is under \textbf{CC BY
4.0}, with code additionally under MIT. Upstream sources keep their
own licenses, listed per source in the release. The WordNet
license permits redistribution with attribution, so the English
definitions and derived items ship in full \citep{fellbaum1998wordnet}.
The Chinese items derive from moedict \citep{moedict2015}, whose
source definitions are CC BY-ND, so verbatim redistribution with
attribution is permitted but adaptations are not. \emph{Two} of our
Chinese transformations are adaptations. One is the answer-avoiding
paraphrase; the other is the OpenCC Traditional\(\to\)Simplified
conversion, since the item builder keys on the converted form. The
release therefore ships \emph{reconstruction tooling rather than
converted text}: the traditional moedict glosses verbatim with
attribution, the pinned upstream \texttt{dict-revised.json} sha256, the exact OpenCC version and \texttt{t2s} configuration, the
paraphrase-regeneration script, and per-item checksums of both
the converted definition and the paraphrase. From these a user rebuilds the
exact prompt text we ran on, and neither adaptation leaves our
machines. Item IDs, answer keys, judgments, and code ship directly.
The SUBTLEX, AoA, concreteness, and CLEARPOND norms ship as per-item
annotation values under their academic-use terms rather than as
redistributed databases.

\end{document}